\documentclass[runningheads]{llncs}

\usepackage{eccv}

\usepackage{eccvabbrv}

\usepackage{graphicx}
\usepackage{booktabs}

\usepackage[accsupp]{axessibility}  % Improves PDF readability for those with disabilities.
\usepackage{multirow}    
\usepackage{float}
\usepackage{adjustbox}
\usepackage{hyperref}

\usepackage{hyperref}

\usepackage{orcidlink}

\begin{document}

% \title{Learning with Virtual Categories for Continual Generalized Category Discovery} 
\title{Virtual Category-Guided Continual Generalized Category Discovery} 

\titlerunning{Abbreviated paper title}

% Second Author\inst{2,3}\orcidlink{1111-2222-3333-4444} \and
% Third Author\inst{3}\orcidlink{2222--3333-4444-5555}}
\author{Jiahui Xiong\inst{1} \and
Qiuxia Lai\inst{2}\textsuperscript{*}\orcidlink{0000-0001-6872-5540} \and
Hongsong Wang\inst{1,3}\textsuperscript{*}\orcidlink{0000-0002-9464-1778}}

\authorrunning{J.~Xiong et al.}

% Springer Heidelberg, Tiergartenstr.~17, 69121 Heidelberg, Germany
% \institute{School of Engineering and Computer Science, Southeast University \and
% State Key Laboratory of Media Convergence and Communication, Communication University of China\\
% \email{\{abc,lncs\}@seu.edu.cn, qxlai@cuc.edu.cn}} % 这个比较全
\institute{
$^1$School of Computer Science and Engineering, Southeast University, Nanjing 210096, China \\
$^2$State Key Laboratory of Media Convergence and Communication, Communication University of China, Beijing 100024, China \\
$^3$Key Laboratory of New Generation Artificial Intelligence Technology and Its Interdisciplinary Applications (Southeast University), Ministry of Education, China \\
\email{\{hongsongwang, jiahuixiong\}@seu.edu.cn, qxlai@cuc.edu.cn}
}

\renewcommand{\thefootnote}{\ensuremath{*}}
\footnotetext[1]{Corresponding authors.}

\maketitle
\begin{abstract}
Continual Generalized Category Discovery (C-GCD) aims to incrementally identify novel categories from sequential unlabeled data while preserving recognition of known classes, which is an essential capability for open-world visual learning. 
A major bottleneck lies in ambiguous unlabeled samples that cannot be confidently assigned to known classes nor reliably grouped as novel ones, making pseudo-labeling brittle and often biasing learning toward familiar categories. 
% In this work, we propose \textit{learning with virtual categories} for C-GCD by adapting Virtual Category Learning (VCL) to the continual setting. 
In this work, we introduce Virtual Category-Guided Continual Generalized Category Discovery by adapting Virtual Category Learning (VCL) to the continual setting. 
Our method identifies uncertain samples and assigns them to temporary virtual categories, enabling safe and informative learning from unlabeled streams without injecting noisy labels, while improving unlabeled data utilization and mitigating prediction bias. To further stabilize discovery across sessions and enhance class separation, we augment VCL with Expanded Neighborhood Contrastive Learning (ENCL), which exploits extended neighborhood relations and an adaptive margin to learn more discriminative and well-separated representations for both old and emerging classes.
Extensive experiments on CIFAR-100, Tiny ImageNet, and ImageNet-100 demonstrate that our approach consistently outperforms state-of-the-art methods, establishing a scalable and effective solution for C-GCD. Code is on: \url{https://github.com/Mrxjh105/VC-CGCD}
% Our code will be released.

% \keywords{Continual Generalized Category Discovery; Continual Learning; Open-World Recognition; Contrastive Learning}
\end{abstract}

\section{Introduction}
\label{sec:intro}

Modern visual recognition has made striking progress in closed-world settings where the label space is fixed and exhaustively defined~\cite{he2016deep,simonyan2014very}. However, real deployments, such as home robots, autonomous driving, and large-scale monitoring, operate in open and evolving environments, where novel categories appear continually and must be incorporated without retraining from scratch~\cite{javed2019meta,wang2026data}. This requirement exposes a fundamental challenge for long-lived learners: they must expand their category set over time while retaining competence on previously learned classes, i.e., the stability–plasticity dilemma. When naively fine-tuned on new data, deep networks are prone to catastrophic forgetting~\cite{mccloskey1989catastrophic, rizve2022openldn, lopez2017gradient}, making continual open-world recognition particularly unstable.

To address emerging categories, \textit{Novel Category Discovery (NCD)}~\cite{han2019learning} studies how to discover unseen classes from unlabeled data given labeled known classes, typically under the assumption that the unlabeled pool contains only novel categories. This assumption rarely holds in realistic streams where old and new categories naturally co-exist. 
\textit{Generalized Category Discovery (GCD)}~\cite{vaze2022generalized} relaxes the setting by allowing unlabeled data to mix known and novel classes, requiring simultaneous recognition of known instances and clustering of novel ones. However, most GCD methods remain batch-oriented and assume access to the full unlabeled pool, making them ill-suited to continuous data arrival. 
\textit{Continual Generalized Category Discovery (C-GCD)}~\cite{zhang2022grow} pushes category discovery into a truly continual regime: after an offline phase trained on labeled known classes, the model receives a sequence of unlabeled sessions that contain both previously seen and newly emerging categories without retaining past data (e.g., due to storage or privacy constraints). Recent progress explores meta-learning~\cite{javed2019meta}, grow-and-merge model updates~\cite{zhang2022grow}, and debiasing strategies~\cite{ma2024happy}, yet practical C-GCD remains challenging.

A central challenge in C-GCD is learning from unlabeled streams under pervasive ambiguity. Online sessions provide no labels, so learning must rely on clustering, neighborhood consistency, or pseudo-labels inferred from the current model. However, unlabeled data are heterogeneous: besides clear known-class instances and discoverable novel-class samples, there exist many ambiguous examples near decision boundaries or in visually overlapping regions. Existing pipelines often force such uncertain samples into either ``known'' or ``novel'' groups via hard pseudo-labels or aggressive clustering. In a continual setting, such early mis-assignments are particularly damaging, as they introduce persistent label noise, amplify confirmation bias, and can compound across sessions when earlier errors cannot be corrected by revisiting data. As a result, the learner may underutilize precisely the hard unlabeled samples that are most informative for carving new decision regions.

% A second difficulty is insufficient feature discrimination between learned and emerging categories, which exacerbates class confusion and forgetting. Contrastive learning is therefore widely used in category discovery, as it can shape representations without requiring labels for all instances by pulling together positive pairs and pushing apart negatives. In NCD/GCD settings, the positive signal is often constructed rather than given: methods may treat nearest neighbors as pseudo-positives~\cite{zhong2021neighborhood}, or build positives via inferred relations over mixed known/new unlabeled data, such as concept-aware alignment or affinity-based supervision~\cite{vaze2022generalized,pu2023dynamic,zhang2023promptcal}. However, in C-GCD these mechanisms can be brittle because representations drift across sessions, making neighbor- or relation-based positive/negative pairs unreliable and yielding unstable assignments~\cite{sun2022opencon}. Consequently, even when unlabeled data are plentiful, the learned embedding can remain under-separated, making both novel class discovery and old-class retention fragile.

\begin{figure*}[t]
\centering
\includegraphics[width=\textwidth]{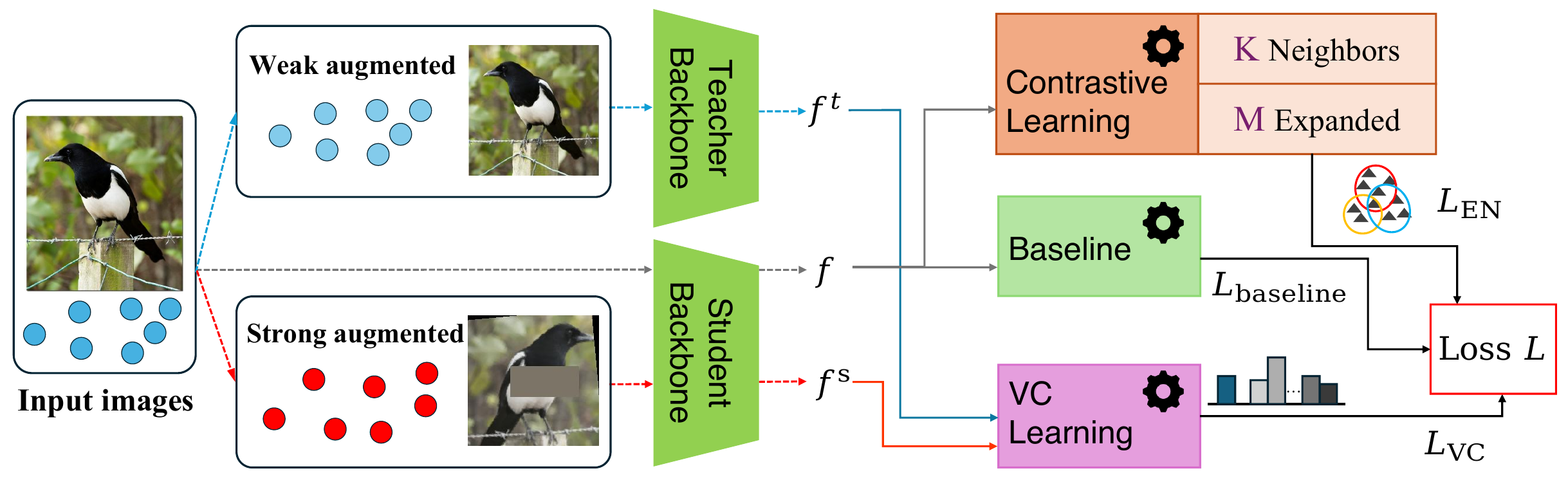}
\caption{Pipeline of the proposed framework. Input images are encoded into shared features, where Virtual Category Learning (VCL) serves as the core mechanism to safely incorporate confusing unlabeled samples through virtual categories, stabilizing learning under evolving category boundaries. Guided by this process, ENCL further improves feature separation, while a baseline objective preserves fundamental recognition performance during continual updates. See \S\ref{VCL} and \S\ref{ENCL} for details.}
\label{overall}
\end{figure*}

This observation motivates the main idea of our paper: learning with virtual categories. Rather than treating every unlabeled sample as if it must belong to one of the current semantic categories, we explicitly provide the model with a mechanism to represent unknown content in a controlled manner. Concretely, we adapt \textit{Virtual Category Learning (VCL)} originally proposed in semi-supervised learning~\cite{chen2024virtual} to the continual generalized discovery setting. VCL assigns uncertain samples to temporary virtual categories, which serve as a buffer that allows the learner to absorb informative structure from ambiguous samples without injecting hard label noise into the known/novel partition. 
% As representations become more reliable over training, samples can be re-aligned away from virtual categories toward stable semantic categories, improving the effectiveness of learning from challenging unlabeled data throughout the continual process. % 这句引发了 R2/mW2, 下面弱化了说法
As representations become more reliable over training, samples are less likely to remain associated with virtual categories and can gradually contribute to stable semantic assignments, improving the effectiveness of learning from challenging unlabeled data throughout the continual process.
To further reduce class confusion and promote separation in the evolving embedding space, inspired by~\cite{banerjee2024amend}, we introduce Expanded Neighborhood Contrastive Learning (ENCL), which leverages neighbors-of-neighbors and an adaptive margin to encourage more discriminative and well-separated representations for both old and emerging classes. Together, VCL and ENCL form a coherent C-GCD framework that learns safely from uncertain unlabeled samples without over-committing to noisy pseudo-labels, while building more discriminative and well-separated representations that mitigate confusion between old and new categories, as summarized in Fig.~\ref{overall}. 

Our main contributions are threefold:
\begin{itemize}
% \item We propose a C-GCD framework that emphasizes \textbf{learning with virtual categories}, addressing ambiguity-safe utilization of unlabeled streams and robust feature separation under representation drift.
\item We introduce the \textbf{Virtual Category-Guided C-GCD framework} that address ambiguity-safe utilization of unlabeled streams and robust feature separation under representation drift.
% \item We adapt \textbf{Virtual Category Learning (VCL)} to C-GCD, assigning uncertain unlabeled samples to temporary virtual categories to improve unlabeled data utilization and reduce error amplification from premature hard assignments.
\item We adapt \textit{Virtual Category Learning (VCL)} to C-GCD by assigning uncertain unlabeled samples to temporary virtual categories, which improves the utilization of unlabeled data and mitigates error amplification caused by premature hard assignments.
% \item We introduce \textbf{Expanded Neighborhood Contrastive Learning (ENCL)} with extended neighborhood sampling and an adaptive margin to capture broader semantic structure and enhance feature discriminability between learned and emerging categories.
\item We introduce \textit{Expanded Neighborhood Contrastive Learning (ENCL)} with extended neighborhood sampling and an adaptive margin to enhance feature discriminability between learned and emerging categories.
\end{itemize}

Extensive experiments on CIFAR-100~\cite{krizhevsky2009learning}, Tiny-ImageNet~\cite{le2015tiny}, and ImageNet-100~\cite{deng2009imagenet} demonstrate that our framework consistently improves accuracy, mitigates forgetting, and enhances novel class discovery over strong C-GCD baselines, validating its effectiveness and practical applicability.

\section{Related Works}
\label{sec:related}

% \noindent\textbf 比 \paragraph 排版更紧凑 【已经全部替换】
\noindent\textbf{Category Discovery.}
\textit{Novel Category Discovery (NCD)} aims to discover previously unseen categories by transferring knowledge from labeled known classes to an unlabeled set that is assumed to contain only novel instances. Early NCD pipelines typically learn a representation on labeled data and then apply unsupervised clustering on unlabeled data for novel class grouping~\cite{han2019learning}. Subsequent studies improve discovery by strengthening representation learning, often leveraging neighborhood- or pseudo-label-based objectives to induce cluster-friendly embeddings under limited supervision~\cite{zhong2021neighborhood, jia2021joint, wang2021progressive, yu2022self, zang2023boosting}.
\textit{Generalized Category Discovery (GCD)} relaxes the key NCD assumption by allowing the unlabeled pool to be a mixture of known and novel classes, requiring the model to recognize known instances while clustering the remaining data into novel categories~\cite{vaze2022generalized}. This formulation better matches realistic deployments, but most existing GCD methods remain \emph{static}, assuming access to the full unlabeled pool and a one-shot training procedure rather than continual updates~\cite{an2023generalized, li2023imbagcd, chiaroni2023parametric, otholt2024guided}.

\noindent\textbf{Continual Generalized Category Discovery.}
\textit{Continual learning (CL)} studies how to learn from sequential data streams without catastrophic forgetting, commonly under supervised task- or class-incremental settings via rehearsal or regularization~\cite{rebuffi2017icarl,lopez2017gradient}. Bringing category discovery into the continual regime yields \textit{Continual Generalized Category Discovery (C-GCD)}, where a model is initialized with labeled known classes and then must learn from a sequence of unlabeled sessions containing both previously seen and emerging categories, typically without storing past data. This setting couples the stability-plasticity tension of CL with the mixed-label-space uncertainty of GCD.
Representative C-GCD methods explore different mechanisms to balance discovery and retention. Grow-and-merge style frameworks expand model capacity or representation diversity and then reconcile conflicts to preserve previously learned knowledge~\cite{zhang2022grow}. Meta-learning-based approaches simulate incremental processes to improve continual adaptation under the mixed known/novel regime~\cite{wu2023metagcd}. Debiasing-oriented frameworks aim to mitigate prediction bias toward known classes and improve robustness under unlabeled continual updates~\cite{ma2024happy}. Despite these advances, C-GCD remains challenging due to ambiguity in fully unlabeled sessions that hinders the reliable utilization of unlabeled data and representation drift that weakens feature discriminability between learned and emerging categories.

\noindent\textbf{Ambiguity and Learning from Unlabeled Data.}
A central challenge in C-GCD is learning from heterogeneous unlabeled streams that include confident known instances, truly novel instances, and ambiguous boundary samples. Many discovery pipelines rely on pseudo-labeling~\cite{yu2022self,wen2023parametric} or clustering-based assignments~\cite{han2019learning,vaze2022generalized}, which can be fragile under ambiguity and may amplify confirmation bias when early errors are reinforced across sessions without the ability to revisit past data. Addressing this requires mechanisms that can utilize ambiguous samples without forcing premature hard commitments.
Recent semi-/self-supervised learning studies propose to handle uncertain samples via auxiliary constructs that reduce label noise and stabilize training. In particular, \emph{virtual category} mechanisms introduce temporary placeholders for uncertain instances to absorb informative structure while avoiding incorrect semantic commitments~\cite{chen2024virtual}. 
Motivated by this challenge, our method introduces Virtual Category Learning (VCL) to handle uncertain unlabeled samples via temporary virtual categories, avoiding premature semantic commitments under continual drift.

\section{Methodology}
\label{sec:method}

\subsection{Preliminaries}
\begin{figure*}[t]
\centering
\includegraphics[width=\textwidth]{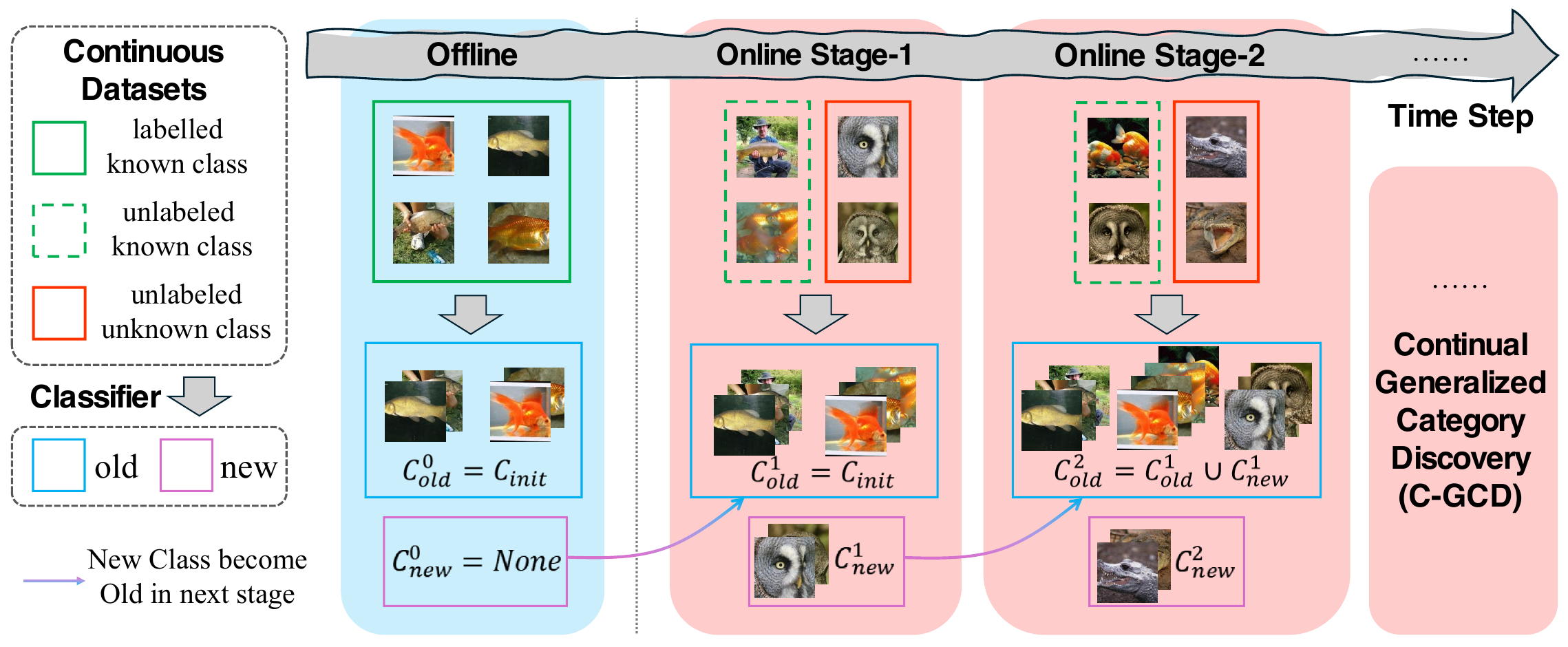}
\caption{C-GCD task setting. The model is initialized with labeled classes $C_\mathrm{init}$ and then receives sequential unlabeled sessions $D^{t}$ containing both old classes $C_\mathrm{old}^{i}$ and novel classes $C_\mathrm{new}^{i}$. The goal is to continually recognize known categories and discover new ones, progressively expanding the known label space across sessions without storing past data. See~\S\ref{sec:problem_formulation} for details.}
\label{task}
\end{figure*}

\noindent\textbf{Problem Formulation.} \label{sec:problem_formulation}
% The C-GCD task can be decoupled into two phases: \textbf{(i)} Offline stage. We train the model on the labeled dataset $D^{0}_{train}=\{(x^{l}_{i},y_{i})\}^{N^{0}}_{i=1}$ at $S_{0}$. The labeled classes in this dataset are $C^{0}_{old}=C_{init}$. Since it is the initialization phase, the unlabeled class is denoted as $C^{0}_{new}=None$.  \textbf{(ii)} Online stage. In each session $S_{t},  t=\{1,2,...,T\}$, the unlabeled dataset $D^{t}_{train}=\{(x^{u}_{i})\}^{N^{t}}_{i=1}$ arrives continuously. We denote the categories in $D^{t}_{train}=\{(x^{u}_{i})\}^{N^{t}}_{i=1}$ as $C^{t}$, which contain $C^{t}_{old}$ and $C^{t}_{new}$. The number of categories is denoted as $K$, including the number of old classes $K^{t}_{old}=|C^{t}_{old}|$ and the number of new classes $K^{t}_{new}=|C^{t}_{new}|$. In this way, we can deduce $K^{t}=K^{t}_{old}+K^{t}_{new}$  from $C^{t}=C^{t}_{old} \cup C^{t}_{new}$. 
As illustrated in Fig.~\ref{task}, C-GCD consists of two phases: an \textbf{offline initialization stage} and an \textbf{online continual discovery stage}. 
\textbf{(i) Offline stage.} At the initial session $S_0$, we train the model on a labeled dataset
$D_{\mathrm{train}}^{0}=\{(\boldsymbol{x}_i^{l}, y_i)\}_{i=1}^{N^{0}}$.
The corresponding label set is denoted by $C_{\mathrm{old}}^{0}=C_{\mathrm{init}}$. Since this is the initialization phase, no novel unlabeled classes are introduced, i.e., $C_{\mathrm{new}}^{0}=\varnothing$.
\textbf{(ii) Online stage.} For each subsequent session $S_t$, $t\in\{1,2,\dots,T\}$, an unlabeled dataset $D_{\mathrm{train}}^{t}=\{\boldsymbol{x}_i^{u}\}_{i=1}^{N^{t}}$ arrives sequentially. Let $C^{t}$ denote the (unknown) category set contained in $D_{\mathrm{train}}^{t}$, which is composed of previously encountered classes and newly emerging classes: $C^{t}=C_{\mathrm{old}}^{t}\cup C_{\mathrm{new}}^{t}$.
We denote the total number of classes in session $t$ by $K^{t}=|C^{t}|$, with $K_{\mathrm{old}}^{t}=|C_{\mathrm{old}}^{t}|$ and $K_{\mathrm{new}}^{t}=|C_{\mathrm{new}}^{t}|$.
Accordingly, $K^{t}=K_{\mathrm{old}}^{t}+K_{\mathrm{new}}^{t}$.

\noindent\textbf{Framework Overview.} \label{sec:overview}
As illustrated in Fig.~\ref{overall}, our method adopts a multi-branch framework for C-GCD and is instantiated on a baseline. For each input image, we generate weakly and strongly augmented views, which are fed into backbone encoders (e.g., ViT or ResNet) to extract feature representations. These features are then processed by two cooperative modules that address complementary challenges in C-GCD.
First, the \textit{Virtual Category Learning (VCL)} module (\S\ref{VCL}) addresses ambiguous unlabeled samples arising from evolving category boundaries. Rather than forcing early semantic assignments, VCL jointly exploits teacher-student features to identify confusing samples and associate them with temporary virtual categories. This allows ambiguous samples to contribute useful training signals while reducing noisy supervision and stabilizing representation learning across sessions.
Second, the \textit{Expanded Neighborhood Contrastive Learning (ENCL)} module (\S\ref{ENCL}) improves feature discriminability by leveraging local neighborhoods with expanded semantic relations, thus improving the separation between previously learned and emerging categories.
% Third, a \textit{baseline learning branch} (\S\ref{loss}) preserves the fundamental discriminative capability of the model. The losses from all branches are optimized jointly in an end-to-end manner. 
Finally, the training objective (\S\ref{loss}) integrates the baseline loss with the proposed VCL and ENCL terms in a unified end-to-end optimization framework. 
Overall, the proposed framework enables the reliable utilization of ambiguous unlabeled data while maintaining robust feature discrimination under continual drift.
% Overall, the proposed method jointly improves ambiguity-safe unlabeled-data utilization and robust feature discrimination, enabling more stable continual category discovery under evolving data distributions.

% \subsection{Virtual Category Discovery}
\subsection{Virtual Category Learning} % 与 introduction 术语一致
\label{VCL}
% TODO: 可以加一段话简短说一下VCL的动机 【已加】
In C-GCD, unlabeled samples contribute to optimization with highly uneven reliability. While confident instances provide relatively stable semantic guidance, a substantial portion of data remains ambiguous due to evolving category boundaries and incremental distribution shifts. Directly imposing hard semantic assignments on such samples often introduces unstable gradients, causing premature feature attraction toward incorrect categories and gradually distorting the embedding space across sessions.
Instead of excluding ambiguous samples or committing them to fixed pseudo labels, we allow uncertain instances to participate in training through \textit{virtual semantic associations}, where their influence is regulated by uncertainty. In this way, ambiguous samples continue to shape representation learning without forcing premature decisions, leading to smoother feature evolution and more stable optimization dynamics throughout continual discovery. 
% Importantly, this treatment does not merely improve sample utilization; it implicitly affects how neighborhood relations emerge during training, since representations are allowed to organize progressively rather than collapsing toward early predictions.
Beyond improving sample utilization, this treatment also influences the emergence of neighborhood structure, since representations can organize progressively instead of collapsing toward early predictions.

% \subsubsection{Confusing Sample}
% \label{sec:csample}

\begin{figure*}[t]
\centering
\includegraphics[width=\textwidth]{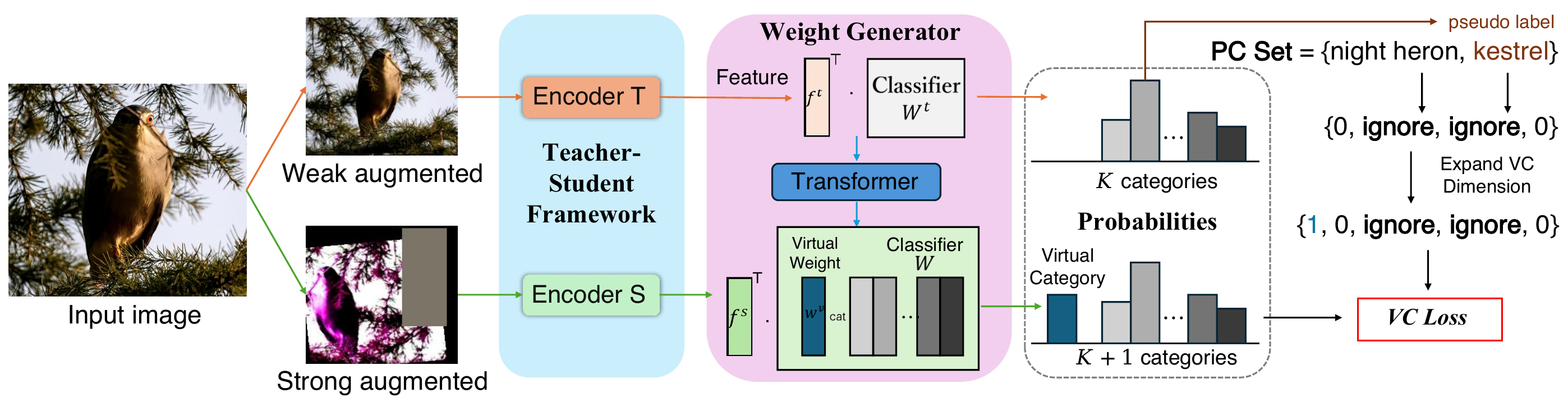}
\caption{Illustration of Virtual Category Learning when handling a confusing sample. Weakly and strongly augmented views are encoded by teacher and student networks to produce prediction scores, from which a potential category set is constructed using prediction competition and teacher--student consistency. Instead of assigning a hard pseudo label, the sample is guided toward a temporary virtual category, allowing it to participate in training without enforcing premature semantic commitment. See \S\ref{VCL}.
}
\label{VCFramework}
\end{figure*}

\noindent\textbf{Confusing Sample.}\label{sec:cs} For clarity, we describe VCL from the perspective of a single sample, while the same procedure applies to mini-batch training. As illustrated in Fig.~\ref{VCFramework}, weakly and strongly augmented views are encoded by the teacher encoder $T$ and the student encoder $S$, producing feature representations for subsequent processing. The teacher classifier computes category probabilities from the projection between the feature vector $({\boldsymbol{f}^{t}})^\top$ and classifier weights $W^{t}$.

Conventional approaches typically use the highest-probability category as a pseudo label for supervision. However, under continual updates, prediction confidence may not reliably reflect semantic correctness, especially when new categories gradually emerge. Treating such predictions as fixed supervision may bias optimization toward transient decision boundaries.
% The VC framework, therefore, interprets uncertainty as an indicator of incomplete semantic separation rather than noise. After obtaining prediction scores, a potential category construction step is introduced to identify whether the sample admits multiple plausible semantic explanations. When more than one candidate category exists, the sample is regarded as a confusing sample. Rather than discarding these samples, VC learning allows them to remain active during training while avoiding directional bias toward any single uncertain category.
VCL instead treats uncertainty as an indicator of incomplete semantic separation rather than  noise. After obtaining prediction scores, a potential-category construction step is performed to determine whether the sample admits multiple plausible semantics. If more than one candidate category exists, the sample is identified as a \textit{confusing sample}. Rather than removing these samples, VCL keeps them active during training while avoiding directional bias toward any single uncertain category.

\begin{figure}[t]
\centering
\includegraphics[width=0.75 \columnwidth]{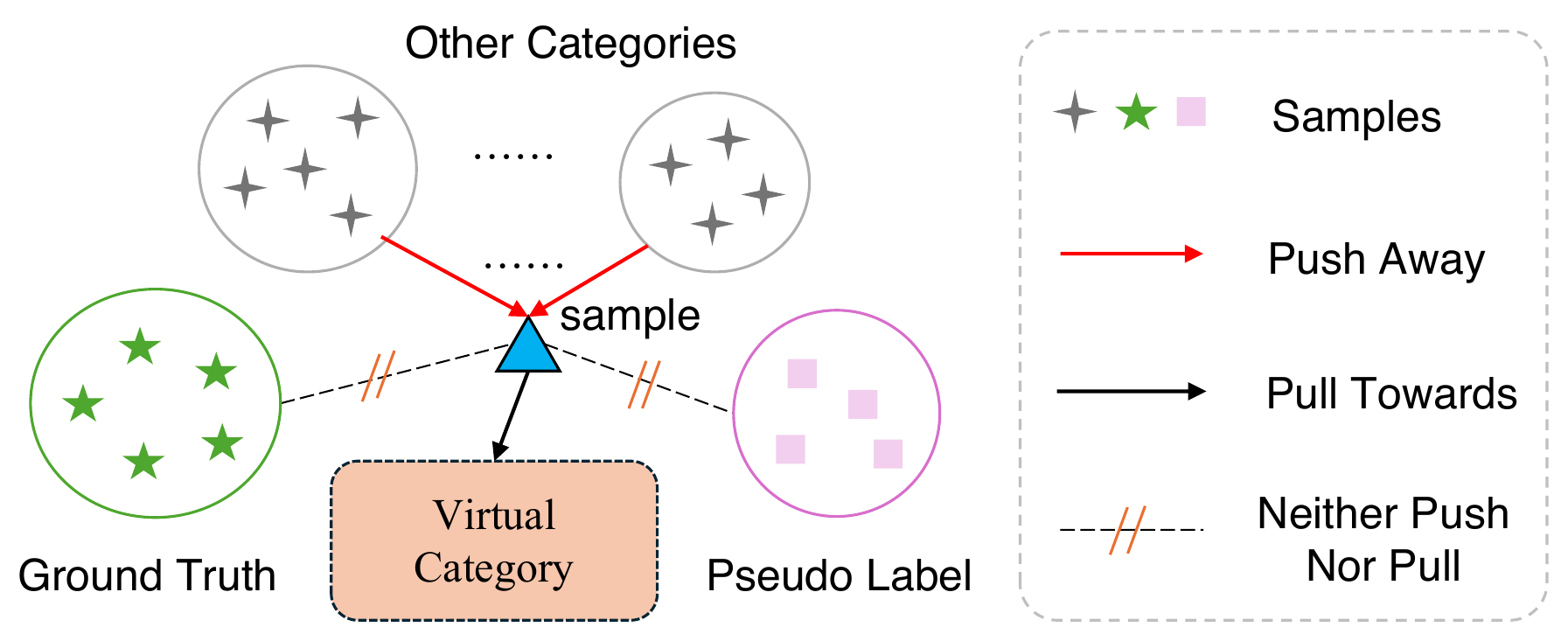}
\caption{Feature-space illustration of Virtual Category Learning. The blue triangle denotes a training sample located near multiple candidate categories, while colored clusters represent different classes and arrows indicate optimization directions. Instead of being strongly attracted to any single category center, the sample is guided toward a virtual category (the dashed rectangular), enabling gradual feature adjustment while avoiding incorrect semantic commitment. See \S\ref{sec:cs} for details.} 
\label{FeatureSpace}
\end{figure}

From a feature-space perspective, as shown in Fig.~\ref{FeatureSpace}, confusing samples should be repelled from clearly incorrect categories, yet aggressively pulling them toward any single candidate risks reinforcing errors. Introducing a virtual category provides a neutral optimization direction, allowing the representation to evolve safely while preserving flexibility for later semantic alignment. In continual learning, such treatment allows ambiguous samples to gradually influence feature organization instead of injecting irreversible mistakes early in training, which in turn reshapes local neighborhood structures over time.

\noindent\textbf{Potential Category Set.}
To realize this idea, we construct a \textit{Potential Category (PC) set}, defined as the collection of candidate categories that remain semantically plausible for an uncertain sample under the current model predictions. Instead of assigning a single pseudo label, the PC set maintains multiple possible category associations, allowing ambiguous samples to contribute to training without enforcing premature decisions. Unlike conventional confidence-threshold filtering, which attempts to eliminate unreliable predictions, the PC formulation preserves ambiguity as informative structural evidence during representation evolution. Maintaining a set of candidates therefore allows optimization to remain flexible while preventing premature semantic commitment.

In practice, the PC set is constructed by jointly considering two complementary uncertainty cues: prediction competition and teacher--student prediction consistency.
The first cue measures whether the most likely categories are difficult to distinguish under the current prediction. Specifically, we compute the relative gap between the top predicted probabilities, $p_\text{1st}$ and $p_\text{2nd}$:
\begin{equation}
    D_\text{top2} = \frac{{p_\text{1st} - p_\text{2nd}}}{{p_\text{2nd}}},
\end{equation}
where a smaller value indicates stronger competition between the top candidate categories.
However, $D_\text{top2}$ is not used as a standalone criterion, since prediction uncertainty may also arise from unstable model responses rather than only from a small top-2 gap.
Therefore, we further compare the predicted labels $y'_{t}$ and $y'_{s}$ from the teacher and student networks. A discrepancy between them indicates prediction inconsistency and provides an additional signal that the sample lies in an uncertain region of the evolving feature space.
The categories suggested by these two cues are combined to form the PC set, and samples with multiple plausible candidates are treated as confusing samples. In this way, low-confidence predictions are not filtered or judged solely by the top-2 score; they can still be identified as confusing when teacher--student predictions are inconsistent.

By integrating these complementary cues, the PC formulation captures uncertainty arising from both local prediction ambiguity and temporal representation variation. As training proceeds, samples within the PC set maintain soft associations with multiple candidate categories, allowing semantic structure to consolidate gradually. This set-based representation thus provides a transitional state between unknown and confidently assigned categories, supporting stable feature organization while category boundaries continue to evolve over sessions.

% 修改说明：先介绍 single sample loss (小l表示)，再介绍 batch-wise loss (L表示)，因为最终的 loss 里面其他 L_EN 之类的 都是 batch-wise 的，所以要统一一下。同时说明是 session-wise loss, 以及为啥 session index t 被省略了。
\noindent\textbf{VC Loss.} 
VCL is applied session-wise in the online stage (i.e., at each $S_t$ to mini-batches sampled from $D_{\mathrm{train}}^{t}$). We first present the loss from the perspective of a single sample in the current session. For notational simplicity, we omit the superscript $t$ in the following equations unless needed. 

Given the VC formulation above, where the PC set is instantiated for C-GCD using prediction competition and teacher--student consistency over old and emerging categories, we introduce a virtual category to absorb ambiguous supervision. Specifically, following~\cite{chen2024virtual}, we extend the classifier weight matrix $W$ with a virtual weight vector $\boldsymbol{w}^{v}$, which is generated by a self-attention transformer layer. The logits are computed as:
% \begin{equation}
%     {\boldsymbol{f}^ \top } \cdot \overbrace {[{\boldsymbol{w}^v},{\boldsymbol{w}^0},...,{\boldsymbol{w}^{K - 1}}]}^{K + 1} = \overbrace {[{l^v},{l^0},...,{l^{K - 1}}]}^{K + 1},
% \end{equation}
\begin{equation}
    \boldsymbol{f}^\top \cdot \left[ \boldsymbol{w}^v, \boldsymbol{w}^0, \dots, \boldsymbol{w}^{K-1} \right] = \left[ l^v, l^0, \dots, l^{K-1} \right],
\end{equation}
where the $l$ represents the logit vector, and the $K$ denotes the number of predicted categories.
For reference, the conventional cross-entropy with pseudo labels obtained through clustering is defined as:
\begin{equation}
    {\ell_\text{CE}} = \log (\sum\nolimits_{i = 0}^K {{e^{{l^i} - {l^{y'}}}}} ).
\end{equation}
With the virtual category, the objective becomes:
\begin{equation} \label{vc_loss_sample}
    {\ell_\text{VC}} = \log (\sum\nolimits_{i = 0,~i \notin \text{PC}}^{K + 2} {{e^{{l^i} - {l^v}}}} ),
\end{equation}
where logits corresponding to categories within the uncertain PC set are excluded from direct supervision. This prevents ambiguous candidates from imposing conflicting gradients while still allowing the sample to influence representation learning through the virtual category.
In practice, at each online session $S_t$, the VCL objective is computed by averaging Eq.~(\ref{vc_loss_sample}) over confusing samples in the current mini-batch $\mathcal{B}$:
\begin{equation}\label{vc_loss}
    \mathcal{L}_{\mathrm{VC}}=\frac{1}{\sum\nolimits_{\boldsymbol{x}_j\in\mathcal{B}} \delta_j}\sum\nolimits_{\boldsymbol{x}_j \in \mathcal{B}} \delta_j \,\ell_{\mathrm{VC}}(\boldsymbol{x}_j),
\end{equation}
where $\delta_j\!\in\!\{0,1\}$ indicates whether $\boldsymbol{x}_j$ is identified as a confusing sample.
As a result, this optimization encourages consistent feature organization without enforcing premature semantic commitments, which provides a more reliable basis for modeling neighborhood relations as categories continue to evolve. 
% Over time, representations shaped under this ambiguity-aware supervision become progressively structured, providing a more reliable basis for modeling neighborhood relations as categories continue to evolve.

% \subsection{Contrastive Learning with Expanded neighborhood}
\subsection{Expanded Neighborhood Contrastive Learning}
% \subsection{Neighborhood Expansion for Continually Evolving Representations}
\label{ENCL}
% TODO: 可以加一段话简短说一下ENCL的动机【已加】
% Virtual Category Learning allows ambiguous samples to participate in training without enforcing premature semantic commitments, which leads to a more flexible but continuously evolving feature space during online learning. As uncertain samples remain active throughout optimization, feature boundaries between old and emerging categories become gradually refined rather than sharply separated at early stages. Under such learning dynamics, conventional neighborhood-based contrastive learning, which relies on strictly local similarity relations, may become overly restrictive and fail to capture broader semantic structures formed during continual updates.
VCL enables ambiguity-safe participation of uncertain samples, improving unlabeled-data utilization but also yielding a feature space that evolves continuously across sessions. To strengthen feature discriminability between learned and emerging categories, 
% a contrastive objective is introduced to regularize embedding geometry. However, conventional neighborhood-based contrastive learning relies on strictly local similarity and can become brittle under continual representation drift.
% To better accommodate this evolving representation space, 
we introduce Expanded Neighborhood Contrastive Learning (ENCL), which extends~\cite{banerjee2024amend} from static GCD to a continual setting to regularize the evolving representation space. Given an image $\boldsymbol{x}_{i}$, traditional neighborhood contrastive learning defines positive samples using nearest neighbors $N(\boldsymbol{x}_i)$, and the loss can be written as:
\begin{equation}  \label{eq5}
    \ell_N =  - \frac{1}{n}\sum\nolimits_{\boldsymbol{q} \in N({\boldsymbol{x}_i})} {\log \frac{{\exp ({\boldsymbol{f}_i} \cdot {\boldsymbol{f}_q}/\tau )}}{{\sum\nolimits_{j = 0,~j \ne i}^{n - 1} {\exp ({\boldsymbol{f}_i} \cdot {\boldsymbol{f}_j}/\tau )} }}},
\end{equation}
where $\boldsymbol{f}_i$ denotes the $l_2$-normalized feature representation of $\boldsymbol{x}_i$, $\tau$ is the temperature parameter, and $n\!=\!\left| {N({\boldsymbol{x}_i})} \right|$.

However, when representations evolve under ambiguity-aware supervision, relying solely on immediate neighbors may overlook semantically related samples that emerge through gradual alignment. We therefore expand the neighborhood by incorporating neighbors-of-neighbors. Specifically, for each neighbor $\boldsymbol{q}\!\in\!N({\boldsymbol{x}_i})$, we retrieve its $m$-nearest neighbors, and construct the expanded neighborhood as:
\begin{equation}
    E_{M}(\boldsymbol{x}_i)=N(\boldsymbol{q});  \quad  \forall \boldsymbol{q} \in N({\boldsymbol{x}_i}).
\end{equation}
This expanded set enlarges the pool of positive samples while preserving locality, enabling the model to capture indirect semantic relations that stabilize contrastive learning across sessions.
Following Eq.~(\ref{eq5}), the expanded neighborhood loss is defined as:
\begin{equation}
    \ell_{EN} =  - \frac{1}{m}\sum\nolimits_{\boldsymbol{q} \in {E_M}({\boldsymbol{x}_i})} {\log \frac{{\exp ({\boldsymbol{f}_i} \cdot {\boldsymbol{f}_q}/\tau )}}{{\sum\nolimits_{j = 0,j \ne i}^{m - 1} {\exp ({\boldsymbol{f}_i} \cdot {\boldsymbol{f}_j}/\tau )} }}},
\end{equation}
where $m\!=\!\left|{{E_M}({\boldsymbol{x}_i})}\right|$ denotes the number of expanded neighbors. Samples appearing multiple times during expansion naturally receive stronger emphasis, reflecting their closer semantic proximity. Combining the original and expanded neighborhoods with a weighting factor $\beta$, the contrastive objective over a mini-batch $\mathcal{B}$ becomes:
\begin{equation} \label{con_loss}
    {\mathcal{L}_\text{EN}} = \sum\nolimits_{i \in \mathcal{B}} {(\ell_N + {\beta}\ell_{EN})}.
\end{equation}
By enlarging neighborhood relations under continually evolving representations, this formulation maintains stable feature structures while allowing gradual separation between previously learned and newly emerging categories.

% \subsection{Unified Optimization Target}
\subsection{Training Objectives}
\label{loss}
% We adopt Happy~\cite{ma2024happy} as our baseline for C-GCD, a debiased framework that mitigates prediction and hardness bias via clustering-guided initialization, soft entropy regularization for novel class discovery, and hardness-aware prototype sampling for alleviating catastrophic forgetting. Happy effectively addresses the inherent bias issues in C-GCD and provides a solid foundation for reliable continual learning. Building on its debiasing mechanisms, our framework further optimizes the utilization of ambiguous unlabeled data and enhances feature discriminability for evolving category spaces.

% The overall training objective follows naturally from the Virtual Category Learning formulation described above. The baseline objective provides stable supervision for previously learned categories, while VCL regulates how uncertain samples participate in optimization by preventing premature semantic commitments. As representations evolve under this learning behavior, the ENCL further preserves local consistency and promotes gradual separation among categories in the embedding space.
% As a result, the final training objective function is:

We train the proposed framework with a unified objective that integrates two task-motivated components: Virtual Category Learning (VCL) for ambiguity-safe utilization of unlabeled samples and Expanded Neighborhood Contrastive Learning (ENCL) for robust feature discrimination under continual drift. In implementation, the framework is instantiated within a C-GCD baseline~\cite{ma2024happy,wu2023metagcd}, whose objective is denoted by $\mathcal{L}_{\text{baseline}}$. 
For brevity, we do not repeat the full decomposition of $\mathcal{L}_{\text{baseline}}$ here and refer readers to~\cite{ma2024happy}.
% We keep the baseline objective unchanged and complement it with $\mathcal{L}_{\mathrm{VC}}$ and $\mathcal{L}_{\mathrm{EN}}$, so that the contribution of the proposed modules can be isolated clearly.
The overall training objective is defined as:
\begin{equation}
    % \mathcal{L} = {\lambda _2}{\underbrace{\mathcal{L}_\text{VC}}_{\text{core}}}  + {\lambda _3}{\mathcal{L}_\text{EN}} +{\mathcal{L}_\text{baseline}},
    \mathcal{L} = \mathcal{L}_\text{VC} + {\lambda_1}{\mathcal{L}_\text{EN}} +\lambda_2{\mathcal{L}_\text{baseline}},
\end{equation}
% where $\lambda_1$ and $\lambda_2$ control the relative influence of representation regularization and baseline loss, respectively. These terms play complementary roles during optimization: the VC loss (Eq.~(\ref{vc_loss})) stabilizes supervision under uncertainty, and the contrastive objective (Eq.~(\ref{con_loss})) maintains coherent feature geometry as the category space expands over sessions. This formulation enables the model to continually incorporate emerging categories while preserving previously acquired knowledge within an end-to-end training procedure.
where the coefficient of $\mathcal{L}_{\text{VC}}$ is fixed to 1 as the reference scale, while $\lambda_1$ and $\lambda_2$ control the relative contributions of the ENCL term and the baseline objective, respectively.
In this formulation, $\mathcal{L}_{\text{VC}}$ regulates how ambiguous samples participate in optimization, $\mathcal{L}_{\text{EN}}$ regularizes neighborhood-based feature geometry to promote separation between previously learned and newly emerging categories, and $\mathcal{L}_{\text{baseline}}$ maintains the basic training behavior of the underlying C-GCD pipeline. These terms are jointly optimized in an end-to-end manner, yielding a unified training framework that improves both unlabeled-data utilization and representation discrimination in C-GCD.

\section{Experiments}
\label{exp}

% In this section, we design experiments in different scenarios to ensure the effectiveness of our work. 
% 加长版凑凑版面
In this section, we evaluate the proposed framework on standard C-GCD benchmarks and examine its effectiveness from three perspectives. We first describe the experimental setups In~\S\ref{sec:exp_setups}, including datasets, evaluation metrics, and implementation details. We then report the main results in \S\ref{sec:main_results} to compare our method with state-of-the-art methods under the C-GCD setting. Finally, in \S\ref{sec:ab_study}, we conduct ablation studies and analysis to isolate the contributions of VCL and ENCL and to further investigate the behavior of the proposed framework.

\subsection{Experimental Setups}\label{sec:exp_setups}
\noindent\textbf{Datasets.} 
We evaluate our method on three widely used benchmarks for category discovery: CIFAR100 (C100)~\cite{krizhevsky2009learning}, ImageNet-100 (IN100)~\cite{deng2009imagenet} and Tiny-ImageNet (Tiny)~\cite{le2015tiny}. Following established C-GCD protocols, each dataset is divided into an offline initialization stage and multiple online continual sessions.

In Stage-0 (offline), the model is trained using labeled data from 50\% of the categories, where each selected class contributes 80\% of its training samples. These classes form the initial known set $C_\mathrm{init}$.

During online stages, the remaining 50\% categories are introduced as novel classes through sequential unlabeled sessions. Each session contains a mixture of previously encountered classes and newly emerging ones. For previously observed categories, a portion of samples naturally reappears in later sessions according to the standard benchmark construction, without storing past data explicitly, thereby preserving the non-rehearsal assumption of C-GCD.

After training on the unlabeled training split $D^{t}_\mathrm{train}$ of each session, evaluation is conducted on the corresponding test split containing both old classes $C^{t}_\mathrm{old}$ and newly introduced classes $C^{t}_\mathrm{new}$.

\noindent\textbf{Evaluation Metrics.} We employ \textbf{accuracy} (\textbf{ACC}) as the primary evaluation metric, reporting performance on:
(i) newly introduced classes (\textit{New}),
(ii) previously learned classes (\textit{Old}), and
(iii) all currently known categories (\textit{All}).

To further analyze continual behavior, we additionally report the \textbf{forgetting} metric $M_{f}$ and the \textbf{discovery} metric $M_{d}$ introduced by \cite{zhang2022grow}, which have been widely used in the continual discovery literature. Specifically,
\begin{equation}
    {M_f} = ACC_\text{known}^0 - \mathop {\min }\limits_{1 \le t \le T} \{ ACC_\text{known}^t\},
\end{equation}
which measures the maximum degradation of previously learned categories across sessions, and
\begin{equation}
    {M_d} = \frac{1}{T}\sum\limits_{1 \le t \le T} {ACC_\text{noval}^t},
\end{equation}
which reflects the sustained discovery capability. Together, these metrics characterize the trade-off between knowledge retention and continual discovery.

\noindent\textbf{Implementation Details.} We adopt a ViT-B/16 \cite{dosovitskiy2020image} backbone pre-trained with DINO initialization. The loss weighting coefficients are set to $\{\beta, \lambda _1, \lambda_2 \}$ as \{0.1, 0.8, 1.0\}. The offline stage is trained for 100 epochs, while each online session is optimized for 30 epochs using a batch size of 128 and a learning rate of 0.01. All experiments are conducted on NVIDIA GeForce RTX 4090D GPUs. Unless otherwise noted, optimization settings follow commonly adopted configurations in C-GCD frameworks to ensure comparable training dynamics across methods.

% \noindent\textbf{Component Analysis.} Our k-NN in ENCL is implemented with GPU-accelerated FAISS for efficiency. Compared with ``w/o ENCL'', our full model increases training time from 338s to 384s per epoch, and GPU memory from 17,936 MiB to 18,824 MiB, indicating a moderate overhead.

% Table generated by Excel2LaTeX from sheet 'main'
\begin{table*}[t]
  \centering
  \caption{Main results of accuracy across offline and 5 online stages on Cifar100(C100), TinyImageNet(Tiny), ImageNet-100(IN100) compared with other works. See \S\ref{anlaysis}.} \label{acc}
   \adjustbox{max width=\textwidth}{
    \begin{tabular}{c|lcccccccccccccccc}
    \toprule
    \multirow{2}[4]{*}{Datasets} & \multicolumn{1}{c}{\multirow{2}[4]{*}{Methods}} & Offline & \multicolumn{3}{c}{Session-1} & \multicolumn{3}{c}{Session-2} & \multicolumn{3}{c}{Session-3} & \multicolumn{3}{c}{Session-4} & \multicolumn{3}{c}{Session-5} \\
\cmidrule{3-18}          &       & All   & All   & Old   & New   & All   & Old   & New   & All   & Old   & New   & All   & Old   & New   & All   & Old   & New \\
    \midrule
          & KMeans & \multicolumn{1}{c|}{66.16 } & 40.27  & 41.76  & \multicolumn{1}{c|}{32.80 } & 37.14  & 38.33  & \multicolumn{1}{c|}{30.00 } & 36.20  & 37.63  & \multicolumn{1}{c|}{26.20 } & 36.66  & 38.30  & \multicolumn{1}{c|}{23.50 } & 35.69  & 36.79  & 25.80  \\
          & VanillaGCD & \multicolumn{1}{c|}{90.82 } & 72.32  & 78.50  & \multicolumn{1}{c|}{41.40 } & 67.04  & 72.50  & \multicolumn{1}{c|}{34.30 } & 57.99  & 62.26  & \multicolumn{1}{c|}{28.10 } & 56.60  & 59.55  & \multicolumn{1}{c|}{33.00 } & 51.36  & 53.70  & 30.30  \\
          & SimGCD & \multicolumn{1}{c|}{90.36 } & 73.37  & 86.44  & \multicolumn{1}{c|}{8.00 } & 62.56  & 72.43  & \multicolumn{1}{c|}{3.30 } & 54.17  & 61.61  & \multicolumn{1}{c|}{2.10 } & 47.62  & 53.37  & \multicolumn{1}{c|}{1.60 } & 43.53  & 47.86  & 4.60  \\
    C100  & FRoST & \multicolumn{1}{c|}{90.36 } & 76.87  & 79.58  & \multicolumn{1}{c|}{63.30 } & 65.31  & 68.88  & \multicolumn{1}{c|}{43.90 } & 58.01  & 61.09  & \multicolumn{1}{c|}{36.50 } & 49.27  & 50.90  & \multicolumn{1}{c|}{36.20 } & 48.03  & 48.17  & 46.80  \\
          & GM    & \multicolumn{1}{c|}{90.36 } & 76.58  & 79.80  & \multicolumn{1}{c|}{60.50 } & 71.10  & 74.52  & \multicolumn{1}{c|}{50.60 } & 63.51  & 68.16  & \multicolumn{1}{c|}{31.00 } & 59.74  & 62.51  & \multicolumn{1}{c|}{37.60 } & 54.11  & 54.74  & 48.40  \\
          & MetaGCD & \multicolumn{1}{c|}{90.82 } & 76.12  & 83.60  & \multicolumn{1}{c|}{38.70 } & 69.40  & 72.82  & \multicolumn{1}{c|}{48.90 } & 61.95  & 65.76  & \multicolumn{1}{c|}{35.30 } & 58.22  & 61.21  & \multicolumn{1}{c|}{34.30 } & 55.78  & 58.47  & 31.60  \\
          & Happy & \multicolumn{1}{c|}{90.36 } & 80.40  & \textbf{85.26} & \multicolumn{1}{c|}{56.10 } & 74.13  & 78.27  & \multicolumn{1}{c|}{49.30 } & 68.23  & 70.86  & \multicolumn{1}{c|}{49.80 } & 62.26  & 63.75  & \multicolumn{1}{c|}{\textbf{50.30 }} & 59.99  & 60.96  & \textbf{51.30} \\
\cmidrule{2-18}          & Ours  & \textbf{90.64} & \textbf{83.22} & 83.56  & \textbf{81.50} & \textbf{76.33} & \textbf{85.26} & \textbf{54.00} & \textbf{70.55} & \textbf{79.74} & \textbf{55.23} & \textbf{64.99} & \textbf{81.94} & 43.80  & \textbf{61.28} & \textbf{62.70} & 48.50  \\
    \midrule
          & KMeans & \multicolumn{1}{c|}{61.70 } & 35.42  & 35.46  & \multicolumn{1}{c|}{35.20 } & 34.99  & 35.75  & \multicolumn{1}{c|}{30.40 } & 34.80  & 36.07  & \multicolumn{1}{c|}{25.90 } & 34.77  & 35.90  & \multicolumn{1}{c|}{24.90 } & 34.62  & 35.63  & 25.50  \\
          & VanillaGCD & \multicolumn{1}{c|}{84.20 } & 55.93  & 58.92  & \multicolumn{1}{c|}{41.00 } & 54.96  & 58.58  & \multicolumn{1}{c|}{33.20 } & 52.82  & 55.74  & \multicolumn{1}{c|}{32.40 } & 48.81  & 51.46  & \multicolumn{1}{c|}{27.60 } & 45.94  & 48.06  & 26.90  \\
          & SimGCD & \multicolumn{1}{c|}{85.86 } & 66.95  & 79.94  & \multicolumn{1}{c|}{2.00 } & 57.81  & 66.98  & \multicolumn{1}{c|}{2.80 } & 52.70  & 59.83  & \multicolumn{1}{c|}{2.77 } & 45.01  & 50.29  & \multicolumn{1}{c|}{2.80 } & 41.59  & 45.79  & 3.80  \\
    Tiny  & FRoST & \multicolumn{1}{c|}{85.86 } & 75.15  & 78.56  & \multicolumn{1}{c|}{58.10 } & 65.64  & 67.83  & \multicolumn{1}{c|}{52.50 } & 51.32  & 54.31  & \multicolumn{1}{c|}{30.40 } & 48.22  & 52.14  & \multicolumn{1}{c|}{16.90 } & 40.15  & 42.73  & 16.90  \\
          & GM    & \multicolumn{1}{c|}{85.86 } & 76.42  & 82.40  & \multicolumn{1}{c|}{46.50 } & 68.87  & 73.82  & \multicolumn{1}{c|}{39.20 } & 58.68  & 63.43  & \multicolumn{1}{c|}{25.40 } & 52.86  & 57.21  & \multicolumn{1}{c|}{18.10 } & 46.90  & 50.62  & 13.40  \\
          & MetaGCD & \multicolumn{1}{c|}{84.20 } & 60.88  & 64.90  & \multicolumn{1}{c|}{40.80 } & 57.20  & 61.03  & \multicolumn{1}{c|}{34.20 } & 54.36  & 57.19  & \multicolumn{1}{c|}{34.60 } & 50.83  & 53.59  & \multicolumn{1}{c|}{28.80 } & 48.14  & 50.16  & 30.00  \\
          & Happy & \multicolumn{1}{c|}{\textbf{85.26}} & 76.67  & \textbf{81.72} & \multicolumn{1}{c|}{51.40 } & 69.50  & \textbf{82.32} & \multicolumn{1}{c|}{37.45 } & 61.78  & \textbf{79.94} & \multicolumn{1}{c|}{31.50 } & 56.17  & \textbf{78.36} & \multicolumn{1}{c|}{28.43 } & 51.99  & \textbf{76.60} & 27.38  \\
\cmidrule{2-18}          & Ours  & 85.15  & \textbf{77.15} & 80.88  & \textbf{58.50} & \textbf{70.54} & 81.38  & \textbf{43.45} & \textbf{63.52} & 78.88  & \textbf{37.93} & \textbf{57.56} & 77.68  & \textbf{32.40} & \textbf{53.25} & 75.98  & \textbf{30.52} \\
    \midrule
          & KMeans & \multicolumn{1}{c|}{85.56 } & 54.90  & 57.04  & \multicolumn{1}{c|}{44.20 } & 54.73  & 56.37  & \multicolumn{1}{c|}{44.90 } & 54.67  & 56.66  & \multicolumn{1}{c|}{40.80 } & 54.63  & 56.25  & \multicolumn{1}{c|}{41.70 } & 53.92  & 56.18  & 33.60  \\
          & VanillaGCD & \multicolumn{1}{c|}{95.96 } & 70.13  & 72.92  & \multicolumn{1}{c|}{56.20 } & 69.37  & 73.47  & \multicolumn{1}{c|}{44.80 } & 68.50  & 70.63  & \multicolumn{1}{c|}{53.60 } & 65.56  & 67.85  & \multicolumn{1}{c|}{47.20 } & 64.54  & 67.44  & 38.40  \\
          & SimGCD & \multicolumn{1}{c|}{96.20 } & 79.67  & 91.68  & \multicolumn{1}{c|}{19.60 } & 70.23  & 78.83  & \multicolumn{1}{c|}{18.60 } & 61.90  & 67.43  & \multicolumn{1}{c|}{23.20 } & 56.67  & 60.92  & \multicolumn{1}{c|}{22.60 } & 52.90  & 56.40  & 21.40  \\
    IN100   & FRoST & \multicolumn{1}{c|}{96.20 } & 87.50  & 92.96  & \multicolumn{1}{c|}{60.20 } & 79.63  & 83.37  & \multicolumn{1}{c|}{57.20 } & 76.78  & 77.00  & \multicolumn{1}{c|}{75.20 } & 66.18  & 68.65  & \multicolumn{1}{c|}{46.40 } & 63.82  & 66.40  & 40.60  \\
          & GM    & \multicolumn{1}{c|}{96.20 } & 89.53  & 95.04  & \multicolumn{1}{c|}{62.00 } & 82.34  & 86.93  & \multicolumn{1}{c|}{54.80 } & 77.97  & 79.17  & \multicolumn{1}{c|}{69.60 } & 72.80  & 74.65  & \multicolumn{1}{c|}{58.00 } & 71.08  & 71.76  & 65.00  \\
          & MetaGCD & \multicolumn{1}{c|}{95.96 } & 75.27  & 78.20  & \multicolumn{1}{c|}{60.60 } & 73.79  & 75.93  & \multicolumn{1}{c|}{54.90 } & 69.35  & 72.20  & \multicolumn{1}{c|}{49.40 } & 67.22  & 70.10  & \multicolumn{1}{c|}{44.20 } & 66.68  & 69.31  & 43.00  \\
          & Happy & \multicolumn{1}{c|}{96.16 } & 91.03  & 95.16  & \multicolumn{1}{c|}{\textbf{70.40 }} & 85.46  & 93.56  & \multicolumn{1}{c|}{65.20 } & 83.07  & 92.92  & \multicolumn{1}{c|}{\textbf{66.67 }} & 76.96  & \textbf{93.24} & \multicolumn{1}{c|}{56.60 } & 74.98  & \textbf{92.40} & 57.56  \\
\cmidrule{2-18}          & Ours  & \textbf{96.21} & \textbf{91.10} & \textbf{95.52} & 69.00  & \textbf{86.31} & \textbf{94.72} & \textbf{65.30} & \textbf{83.33} & \textbf{93.96} & 65.60  & \textbf{78.60} & 92.24  & \textbf{61.55} & \textbf{76.72} & 91.88  & \textbf{61.55} \\
    \bottomrule
    \end{tabular}
    }
  \label{tab:addlabel}%
\end{table*}

% Table generated by Excel2LaTeX from sheet 'Mf&Md'
\begin{table}[t]
  \centering
  \caption{Comparison of forgetting $M_f$ and discovery $M_d$ metrics which demonstrate improved stability and sustained discovery capability of the proposed method. See \S\ref{anlaysis}.} 
  % \adjustbox{max width=0.6 \linewidth}{
  \resizebox{0.7\textwidth}{!}{
  \setlength\tabcolsep{8.0pt}
  \renewcommand\arraystretch{1.2} % 用这3行命令，调节表格格式更灵活
    \begin{tabular}{lrrrr}
    \toprule
    \multicolumn{1}{c}{\multirow{2}[4]{*}{Methods}} & \multicolumn{2}{c}{C100} & \multicolumn{2}{c}{Tiny} \\
\cmidrule{2-5}          & \multicolumn{1}{c}{$M_f~(\downarrow)$} & \multicolumn{1}{c}{$M_d~(\uparrow)$} & \multicolumn{1}{c}{$M_f~(\downarrow)$} & \multicolumn{1}{c}{$M_d~(\uparrow)$} \\
    \midrule
    VanillaGCD\cite{vaze2022generalized} & 37.12  & 33.42  & 36.14  & 32.22  \\
    FRoST\cite{roy2022class} & 42.19  & 45.34  & 43.13  & 34.96  \\
    MetaGCD\cite{wu2023metagcd} & 32.35  & 37.76  & 34.04  & 33.68  \\
    Happy\cite{ma2024happy} & 29.40  & 51.36  & \textbf{8.66} & 35.23  \\
    \midrule
    Ours  & \textbf{27.94} & \textbf{56.61} & 9.17 & \textbf{40.56} \\
    \bottomrule
    \end{tabular}
    }
  \label{f&d}
\end{table}%

% Table generated by Excel2LaTeX from sheet '10stage'
\begin{table*}[t]
  \centering
  \caption{Average ``All'' accuracy across 10 continual sessions, evaluating long-term performance under extended online updates. Our method shows improved robustness to continual distribution shifts and sustained category discovery ability. See \S\ref{anlaysis}.} 
  % \adjustbox{max width=\textwidth}{
  \resizebox{\textwidth}{!}{
  \setlength\tabcolsep{3.2pt}
  \renewcommand\arraystretch{1.2}
    \begin{tabular}{rlccccccccccc}
    \toprule
    \multicolumn{1}{l}{Data} & Methods & 1     & 2     & 3     & 4     & 5     & 6     & 7     & 8     & 9     & 10    & Avg \\
    \midrule
    \multirow{2}{*}{C100} & Happy & 85.20  & 81.63  & 77.94  & 74.57  & 69.16  & 66.25  & 62.34  & 59.72  & 54.49  & 52.89  & 68.42  \\
          & Ours  & 85.60  & 83.35  & 80.51  & 75.24  & 71.17  & 68.23  & 63.73  & 61.03  & 59.51  & 54.87  & \textbf{70.32} \\
    \midrule
    \multirow{2}{*}{Tiny} & Happy & 79.35  & 75.63  & 71.28  & 66.83  & 62.53  & 60.05  & 55.09  & 52.01  & 49.79  & 45.77  & 61.83  \\
          & Ours  & 79.25  & 76.13  & 72.28  & 68.21  & 64.25  & 61.74  & 55.95  & 53.28  & 50.60  & 47.42  & \textbf{62.91} \\
    \bottomrule
    \end{tabular}
    }
  \label{10stage}
\end{table*}

\subsection{Main Results}~\label{sec:main_results}
\label{anlaysis}
We report the ACC results of different methods in Table \ref{acc}. In general, our work achieves the best performance in terms of ACC in most cases, which demonstrates the effectiveness of our work for the C-GCD task. It is observed that, compared with previous methods: 1) KMeans\cite{mcqueen1967some} on pre-trained features, 2) GCD methods: VanillaGCD\cite{vaze2022generalized}, SimGCD\cite{wen2023parametric}, 3) novel C-GCD works: FRoST\cite{roy2022class}, GM\cite{zhang2022grow}, MetaGCD\cite{wu2023metagcd}, Happy\cite{ma2024happy}. Our method maintains higher accuracy throughout the continual process while avoiding rapid degradation in later sessions. A clear improvement is observed on newly emerging categories. In contrast to methods that rely on hard pseudo-labeling or fixed neighborhood assumptions, ambiguity-aware supervision allows uncertain samples to contribute without enforcing premature semantic assignments. As a result, newly introduced classes are discovered more reliably, leading to improved New accuracy across most stages. For example, our approach improves the overall performance by 1.29\% on C100, 1.26\% on Tiny, and 1.74\% on IN100, with particularly noticeable gains in new categories in later sessions.

We further evaluate forgetting and discovery capability in Table \ref{f&d}. The $M_d$ values of our method all achieve the best results on both C100 and Tiny by $10.2 \sim 15.2\%$. Meanwhile, competitive $M_f$ values show that the model preserves knowledge of old classes while incorporating new categories. These results suggest that preventing premature commitments helps reduce confirmation bias and stabilizes learning across sessions.

To examine long-term behavior, we extend the experiment to 10 stages, as shown in Table \ref{10stage}. The proposed method maintains a higher average accuracy over extended continual updates.

\subsection{Ablation Study and Analysis}\label{sec:ab_study}

% \noindent\textbf{Ablation Study.}
\noindent\textbf{Component Analysis.}
\label{ab_an}
To better understand the performance gains observed in the main results, we report ablation results in Table \ref{ablation} by removing the VCL and ENCL on CIFAR-100 and Tiny-ImageNet. Removing either component consistently degrades performance, indicating that stable continual discovery relies on both ambiguity regulation and representation refinement during training. Without ambiguity-aware supervision, uncertain samples introduce noisier updates, while removing neighborhood expansion weakens the gradual organization of features across sessions. When both components are enabled, the model achieves the best performance across \textit{All}, \textit{Old}, and \textit{New} categories, suggesting that the utilization of confusing samples and evolving neighborhood relations jointly supports more stable category formation over time.

% Table generated by Excel2LaTeX from sheet '消融'
\begin{table}[t]
  \centering
  \caption{Ablation study on the main modules, reporting the average accuracy (ACC) across all stages for C100 and Tiny. The results highlight the contributions of VCL and contrastive neighborhood refinement to stable C-GCD. See \S\ref{ab_an} for details.} 
  \adjustbox{max width=0.6\linewidth}{
    \begin{tabular}{cc|ccc|ccc}
    \toprule
          &       &       & CIFAR-100 &       &       & Tiny-ImageNet &  \\
    \midrule
    VC    & EN & All   & Old   & New   & All   & Old   & New \\
    \midrule
    ×     & ×     & 69.00  & 71.82  & 51.36  & 63.22  & 79.79  & 35.23  \\
    \checkmark     & ×     & 70.04  & 80.06  & 50.45  & 63.13  & 79.92  & 35.14  \\
    ×     & \checkmark     & 69.57  & 80.96  & 51.10  & 64.22  & 78.93  & 40.22  \\
    \checkmark     & \checkmark     & \textbf{71.27} & 78.64  & 56.61  & \textbf{64.40} & 78.96  & 40.56  \\
    \bottomrule
    \end{tabular}
    }
  \label{ablation}
\end{table}

% \subsection{Extended Analysis}
% We design two additional metrics to further analyze the learning behavior behind the performance improvements.

\begin{figure}[t]
\centering
\begin{minipage}{0.48\columnwidth}
\centering
\includegraphics[width=\linewidth]{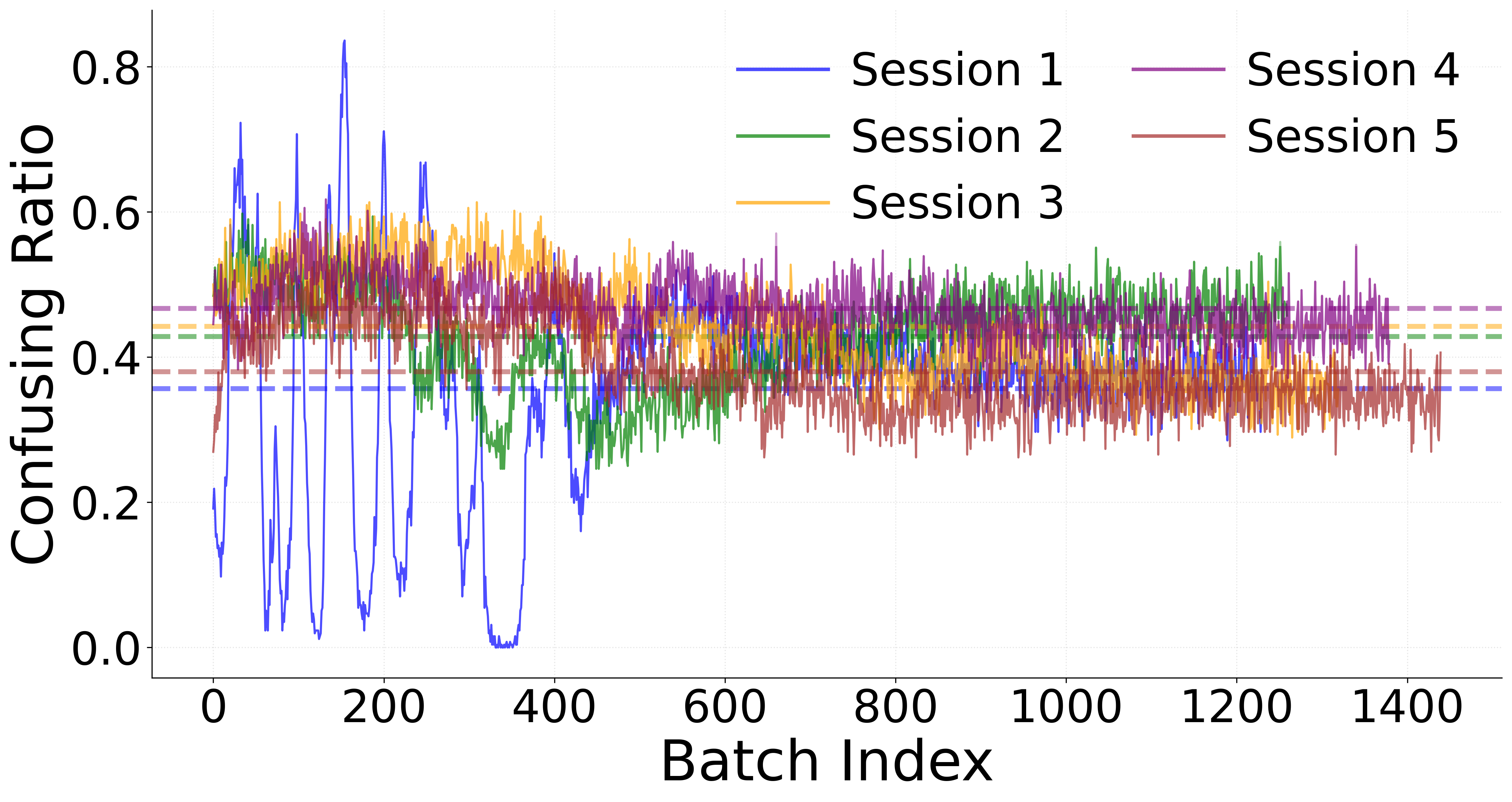}
\caption{Ratio of confusing samples used in training across sessions. This metric indicates how uncertain samples are handled during continual updates, reflecting the model’s ability to utilize ambiguous data without forcing premature decisions.}
\label{ConfusingRatio}
\end{minipage}
\hfill  % 或 \hspace{0.04\columnwidth}
\begin{minipage}{0.48\columnwidth}
\centering
\includegraphics[width=\linewidth]{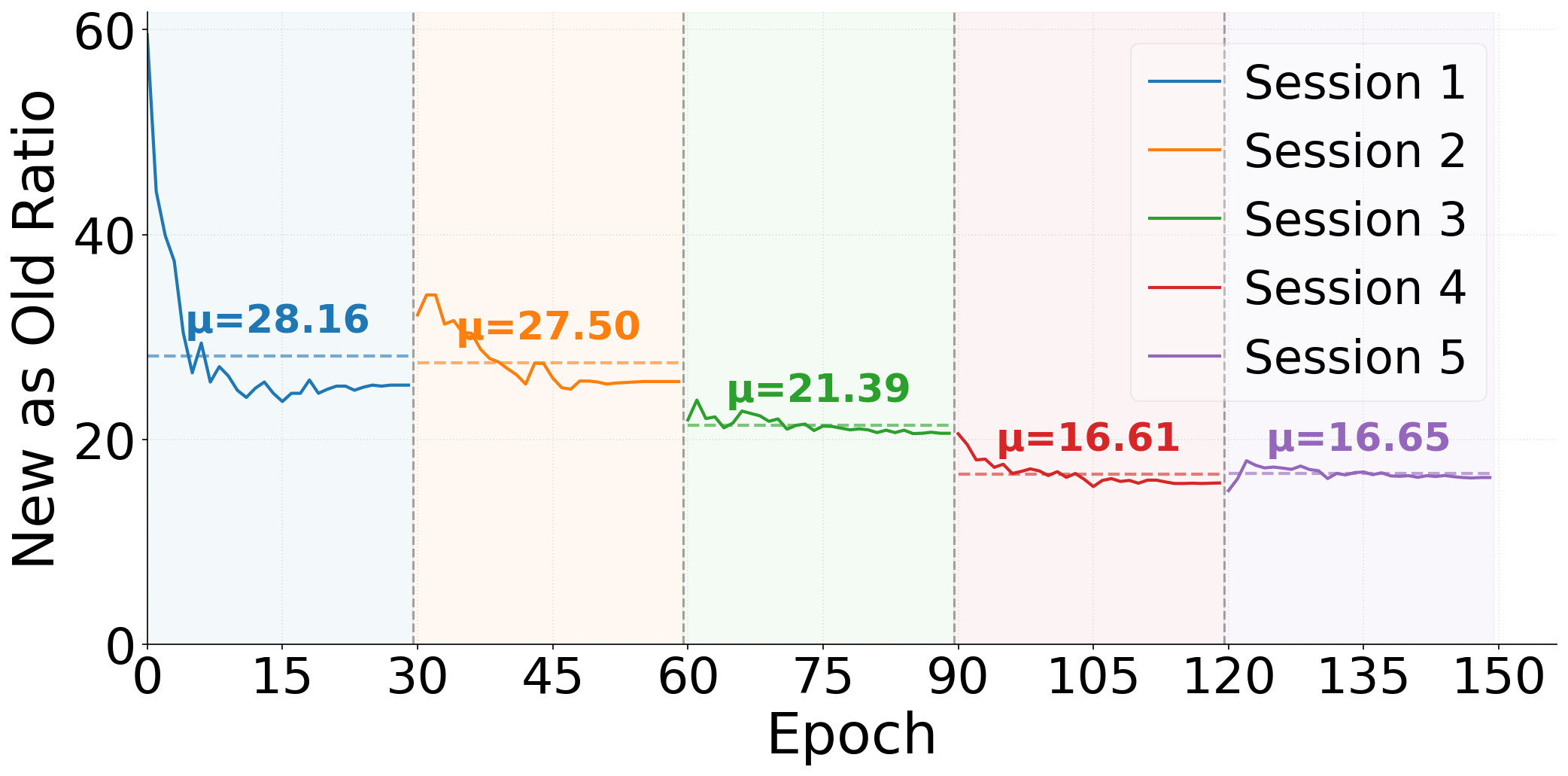}
\caption{Proportion of novel-class samples misclassified as old classes in each session. This metric measures the separability between old and emerging categories, showing how feature organization improves over time. See \S\ref{sec:ab_study} for details.}
\label{NewAsOld}
\end{minipage}
\end{figure}

% \noindent\textbf{Confusing Ratio.} The proportion of confusing samples among all training data indicates how uncertainty is handled throughout training. As shown in Fig.~\ref{ConfusingRatio}, the ratio fluctuates at the beginning of Session 1 (the blue curve) and gradually stabilizes in later sessions. Similar but weaker variations appear afterward, suggesting that confusing samples play a larger role during early adaptation and become progressively regulated as representations mature. Although the ratio slightly decreases in later stages, it remains within a stable range, indicating balanced utilization of uncertain samples throughout continual updates.
\noindent\textbf{Confusing Ratio.} The proportion of confusing samples among all training data indicates how uncertainty is handled throughout training. Since confusing samples are identified by ambiguity rather than compact class membership, this ratio provides an indirect batch-level view of how VCL buffers and regulates these samples over time. As shown in Fig.~\ref{ConfusingRatio}, the ratio fluctuates at the beginning of Session 1 (the blue curve) and gradually stabilizes in later sessions. Similar but weaker variations appear afterward, suggesting that confusing samples play a larger role during early adaptation and become progressively regulated as representations mature. Although the ratio slightly decreases in later stages, it remains within a stable range, indicating balanced utilization of uncertain samples throughout continual updates.

\noindent\textbf{New as Old Ratio.}
We further measure the proportion of novel-class samples misclassified as old classes to evaluate category separability. Fig.~\ref{NewAsOld} shows that this ratio consistently decreases within each session, while the session-wise mean gradually stabilizes. This trend suggests that feature organization improves over time, reducing bias toward previously learned categories and enabling clearer separation between old and emerging classes.

% \subsection{Hyperparameter Sensitivity Analysis} 
\noindent\textbf{Hyperparameter Sensitivity Analysis.} 
All hyperparameters are selected via validation and kept unchanged across all experimental settings to ensure fair comparison. To examine the robustness of our framework and justify our default parameter choices, we conduct sensitivity studies on two core components: the ambiguity threshold for PC set construction and the neighborhood size adopted in ENCL. As reported in Table \ref{tab:ablation_params}, our default settings (threshold 0.15, neighborhood size 3) consistently achieve the best average accuracy among nearby values. We note that the PC set retains samples satisfying both the top-2 prediction ambiguity and teacher–student consistency criteria, and the virtual-category weights are learned automatically during training instead of being manually tuned.
\begin{table}[t]
  \centering
  \caption{Sensitivity analysis of ambiguity threshold and neighborhood size on C100.}
  \label{tab:ablation_params}
  \vspace{-6pt}
    \setlength\tabcolsep{2.0pt}
    \renewcommand\arraystretch{1.0}
  \resizebox{0.9\linewidth}{!}{%
  \begin{tabular}{c|ccccc||c|ccc}
    \hline\noalign{\hrule height 0.1pt}
    Threshold & 0.07  & 0.1   & \textbf{0.15} & 0.2   & 0.3   & Neighborhood Size & 2     & \textbf{3} & 4 \\
    \hline
    Avg   & 68.63 & 70.53 & \textbf{71.27} & 70.12 & 68.42 & Avg   & 70.52 & \textbf{71.27} & 69.09 \\
    \hline
  \end{tabular}%
  }
  % \vspace{-6pt}
\end{table}

\noindent\textbf{Computational Efficiency.} Our k-NN in ENCL is implemented with GPU-accelerated FAISS for efficiency. As summarized in Table~\ref{tab:efficiency}, compared with ``w/o ENCL'', our full model increases training time from 338s to 384s per epoch, and GPU memory from 17,936 MiB to 18,824 MiB, indicating a moderate overhead.
% 上方正文结束后，压缩表格顶部与上文的距离
% \vspace{-8pt}
\begin{table}[t]
\centering
\setlength{\tabcolsep}{6pt}
\renewcommand{\arraystretch}{0.9}
% 压缩标题与表格线之间的距离
\setlength{\abovecaptionskip}{4pt}   % 标题上方留白
\setlength{\belowcaptionskip}{2pt}   % 标题下方与表格线之间留白
\caption{Computational overhead of ENCL on C100 with ViT-B/16.}
\label{tab:efficiency}
\begin{tabular}{lrr}
\toprule
Metric                  & w/o ENCL & Full model \\
\midrule
Training Time (s/epoch) & 338      & 384        \\
GPU Memory (MiB)        & 17,936   & 18,824     \\
\bottomrule
\end{tabular}
\vspace{-6pt}
\end{table}
% 压缩表格底部与下文的距离

\vspace{-6pt}
\section{Conclusion}
\label{conclusion}
% \vspace{-6pt}
In this paper, we present an ambiguity-aware framework for Continual Generalized Category Discovery that improves learning from unlabeled streams under continual distribution shifts. Instead of forcing uncertain samples into premature semantic assignments, our approach introduces Virtual Category Learning (VCL) as a principled mechanism to safely absorb ambiguous data through temporary virtual categories, enabling stable knowledge accumulation while mitigating error propagation across sessions. By allowing uncertain samples to contribute to representation learning without injecting noisy supervision, the proposed framework improves data utilization and alleviates prediction bias toward previously learned classes. Built upon this foundation, an expanded neighborhood contrastive strategy further refines the learned embedding space, promoting clearer separation between evolving categories and enhancing discovery stability over time. Extensive experiments across multiple benchmarks demonstrate consistent gains in overall recognition, novel class discovery, and forgetting mitigation, validating the effectiveness of learning with virtual categories for continual open-world recognition. Our results suggest that explicitly modeling uncertainty is a key step toward scalable and reliable continual category discovery systems.

% \vspace{-6pt}
\section*{Acknowledge}
\label{acknowledge}
% \vspace{-6pt}
This work was supported by National Science Foundation of China (62302093, 62306292, 52441503), Jiangsu Province Natural Science Fund (BK20230833) and the Open Research Fund of the State Key Laboratory of Multimodal Artificial Intelligence Systems under Grant E5SP060116. We thank the Big Data Computing Center of Southeast University for providing the facility support on the numerical calculations. 

\bibliographystyle{splncs04}
\bibliography{main}

@String(AAAI  = {AAAI})

@article{krizhevsky2009learning,
  title={Learning multiple layers of features from tiny images},
  author={Krizhevsky, Alex and Hinton, Geoffrey and others},
  journal={Technical report, University of Toronto},
  year={2009},
  publisher={Toronto, ON, Canada}
}

@article{le2015tiny,
  title={Tiny imagenet visual recognition challenge},
  author={Le, Yann and Yang, Xuan},
  journal={CS 231N},
  volume={7},
  number={7},
  pages={3},
  year={2015}
}

@inproceedings{deng2009imagenet,
  title={Imagenet: A large-scale hierarchical image database},
  author={Deng, Jia and Dong, Wei and Socher, Richard and Li, Li-Jia and Li, Kai and Fei-Fei, Li},
  booktitle={2009 IEEE Conference on Computer Vision and Pattern Recognition},
  pages={248--255},
  year={2009},
  organization={Ieee}
}

@article{ma2024happy,
  title={Happy: A debiased learning framework for continual generalized category discovery},
  author={Ma, Shijie and Zhu, Fei and Zhong, Zhun and Liu, Wenzhuo and Zhang, Xu-Yao and Liu, Cheng-Lin},
  journal={Advances in Neural Information Processing Systems},
  volume={37},
  pages={50850--50875},
  year={2024}
}

@article{zhang2022grow,
  title={Grow and merge: A unified framework for continuous categories discovery},
  author={Zhang, Xinwei and Jiang, Jianwen and Feng, Yutong and Wu, Zhi-Fan and Zhao, Xibin and Wan, Hai and Tang, Mingqian and Jin, Rong and Gao, Yue},
  journal={Advances in Neural Information Processing Systems},
  volume={35},
  pages={27455--27468},
  year={2022}
}

@article{dosovitskiy2020image,
  title={An image is worth 16x16 words: Transformers for image recognition at scale},
  author={Dosovitskiy, Alexey and Beyer, Lucas and Kolesnikov, Alexander and Weissenborn, Dirk and Zhai, Xiaohua and Unterthiner, Thomas and Dehghani, Mostafa and Minderer, Matthias and Heigold, Georg and Gelly, Sylvain and others},
  journal={arXiv preprint arXiv:2010.11929},
  year={2020}
}

@inproceedings{han2019learning,
  title={Learning to discover novel visual categories via deep transfer clustering},
  author={Han, Kai and Vedaldi, Andrea and Zisserman, Andrew},
  booktitle={Proceedings of the IEEE/CVF International Conference on Computer Vision},
  pages={8401--8409},
  year={2019}
}

@inproceedings{vaze2022generalized,
  title={Generalized category discovery},
  author={Vaze, Sagar and Han, Kai and Vedaldi, Andrea and Zisserman, Andrew},
  booktitle={Proceedings of the IEEE/CVF Conference on Computer Vision and Pattern Recognition},
  pages={7492--7501},
  year={2022}
}

@inproceedings{rebuffi2017icarl,
  title={icarl: Incremental classifier and representation learning},
  author={Rebuffi, Sylvestre-Alvise and Kolesnikov, Alexander and Sperl, Georg and Lampert, Christoph H},
  booktitle={Proceedings of the IEEE/CVF Conference on Computer Vision and Pattern Recognition},
  pages={2001--2010},
  year={2017}
}

@article{lopez2017gradient,
  title={Gradient episodic memory for continual learning},
  author={Lopez-Paz, David and Ranzato, Marc'Aurelio},
  journal={Advances in Neural Information Processing Systems},
  volume={30},
  year={2017}
}

@inproceedings{banerjee2024amend,
  title={Amend: Adaptive margin and expanded neighborhood for efficient generalized category discovery},
  author={Banerjee, Anwesha and Kallooriyakath, Liyana Sahir and Biswas, Soma},
  booktitle={Proceedings of the IEEE/CVF Winter Conference on Applications of Computer Vision},
  pages={2101--2110},
  year={2024}
}

@inproceedings{wu2023metagcd,
  title={Metagcd: Learning to continually learn in generalized category discovery},
  author={Wu, Yanan and Chi, Zhixiang and Wang, Yang and Feng, Songhe},
  booktitle={Proceedings of the IEEE/CVF International Conference on Computer Vision},
  pages={1655--1665},
  year={2023}
}

@inproceedings{roy2022class,
  title={Class-incremental novel class discovery},
  author={Roy, Subhankar and Liu, Mingxuan and Zhong, Zhun and Sebe, Nicu and Ricci, Elisa},
  booktitle={European Conference on Computer Vision},
  pages={317--333},
  year={2022},
  organization={Springer}
}

@inproceedings{mcqueen1967some,
  title={Some methods of classification and analysis of multivariate observations},
  author={McQueen, James B},
  booktitle={Proc. of 5th Berkeley Symposium on Math. Stat. and Prob.},
  pages={281--297},
  year={1967}
}

@inproceedings{wen2023parametric,
  title={Parametric classification for generalized category discovery: A baseline study},
  author={Wen, Xin and Zhao, Bingchen and Qi, Xiaojuan},
  booktitle={Proceedings of the IEEE/CVF International Conference on Computer Vision},
  pages={16590--16600},
  year={2023}
}

@inproceedings{he2016deep,
  title={Deep residual learning for image recognition},
  author={He, Kaiming and Zhang, Xiangyu and Ren, Shaoqing and Sun, Jian},
  booktitle={Proceedings of the IEEE/CVF Conference on Computer Vision and Pattern Recognition},
  pages={770--778},
  year={2016}
}

@article{simonyan2014very,
  title={Very deep convolutional networks for large-scale image recognition},
  author={Simonyan, Karen and Zisserman, Andrew},
  journal={arXiv preprint arXiv:1409.1556},
  year={2014}
}

@article{javed2019meta,
  title={Meta-learning representations for continual learning},
  author={Javed, Khurram and White, Martha},
  journal={Advances in Neural Information Processing Systems},
  volume={32},
  year={2019}
}

@incollection{mccloskey1989catastrophic,
  title={Catastrophic interference in connectionist networks: The sequential learning problem},
  author={McCloskey, Michael and Cohen, Neal J},
  booktitle={Psychology of Learning and Motivation},
  volume={24},
  pages={109--165},
  year={1989}
}

@article{chen2024virtual,
  title={Virtual category learning: A semi-supervised learning method for dense prediction with extremely limited labels},
  author={Chen, Changrui and Han, Jungong and Debattista, Kurt},
  journal={IEEE Transactions on Pattern Analysis and Machine Intelligence},
  volume={46},
  number={8},
  pages={5595--5611},
  year={2024}
}

@inproceedings{zhong2021neighborhood,
  title={Neighborhood contrastive learning for novel class discovery},
  author={Zhong, Zhun and Fini, Enrico and Roy, Subhankar and Luo, Zhiming and Ricci, Elisa and Sebe, Nicu},
  booktitle={Proceedings of the IEEE/CVF Conference on Computer Vision and Pattern Recognition},
  pages={10867--10875},
  year={2021}
}

@inproceedings{jia2021joint,
  title={Joint representation learning and novel category discovery on single-and multi-modal data},
  author={Jia, Xuhui and Han, Kai and Zhu, Yukun and Green, Bradley},
  booktitle={Proceedings of the IEEE/CVF International Conference on Computer Vision},
  pages={610--619},
  year={2021}
}

@article{wang2021progressive,
  title={Progressive self-supervised clustering with novel category discovery},
  author={Wang, Jingyu and Ma, Zhenyu and Nie, Feiping and Li, Xuelong},
  journal={IEEE Transactions on Cybernetics},
  volume={52},
  number={10},
  pages={10393--10406},
  year={2021},
  publisher={IEEE}
}

@inproceedings{yu2022self,
  title={Self-labeling framework for novel category discovery over domains},
  author={Yu, Qing and Ikami, Daiki and Irie, Go and Aizawa, Kiyoharu},
  booktitle={Proceedings of the AAAI Conference on Artificial Intelligence},
  volume={36},
  number={3},
  pages={3161--3169},
  year={2022}
}

@inproceedings{zang2023boosting,
  title={Boosting novel category discovery over domains with soft contrastive learning and all in one classifier},
  author={Zang, Zelin and Shang, Lei and Yang, Senqiao and Wang, Fei and Sun, Baigui and Xie, Xuansong and Li, Stan Z},
  booktitle={Proceedings of the IEEE/CVF International Conference on Computer Vision},
  pages={11858--11867},
  year={2023}
}

@inproceedings{an2023generalized,
  title={Generalized category discovery with decoupled prototypical network},
  author={An, Wenbin and Tian, Feng and Zheng, Qinghua and Ding, Wei and Wang, QianYing and Chen, Ping},
  booktitle={Proceedings of the AAAI Conference on Artificial Intelligence},
  volume={37},
  number={11},
  pages={12527--12535},
  year={2023}
}

@article{li2023imbagcd,
  title={ImbaGCD: Imbalanced generalized category discovery},
  author={Li, Ziyun and Dai, Ben and Simsek, Furkan and Meinel, Christoph and Yang, Haojin},
  journal={arXiv preprint arXiv:2401.05353},
  year={2023}
}

@inproceedings{chiaroni2023parametric,
  title={Parametric information maximization for generalized category discovery},
  author={Chiaroni, Florent and Dolz, Jose and Masud, Ziko Imtiaz and Mitiche, Amar and Ben Ayed, Ismail},
  booktitle={Proceedings of the IEEE/CVF International Conference on Computer Vision},
  pages={1729--1739},
  year={2023}
}

@inproceedings{otholt2024guided,
  title={Guided cluster aggregation: A hierarchical approach to generalized category discovery},
  author={Otholt, Jona and Meinel, Christoph and Yang, Haojin},
  booktitle={Proceedings of the IEEE/CVF Winter Conference on Applications of Computer Vision},
  pages={2618--2627},
  year={2024}
}

@inproceedings{rizve2022openldn,
  title={Openldn: Learning to discover novel classes for open-world semi-supervised learning},
  author={Rizve, Mamshad Nayeem and Kardan, Navid and Khan, Salman and Shahbaz Khan, Fahad and Shah, Mubarak},
  booktitle={European Conference on Computer Vision},
  pages={382--401},
  year={2022},
  organization={Springer}
}

@article{zhao2025multi,
  title={Multi-view contrastive learning with maximal mutual information for continual generalized category discovery},
  author={Zhao, Zihao and Li, Xiao and Chang, Zhonghao and Hu, Ningge},
  journal={Expert Systems with Applications},
  volume={266},
  pages={125994},
  year={2025},
  publisher={Elsevier}
}

@inproceedings{hao2025tree,
  title={Tree of Prompts: Aligning Hierarchical Visual Prior for Continual Generalized Category Discovery},
  author={Hao, Yiqing and Huang, Yangru and Jin, Yi and Wang, Tao and Li, Yidong and Cen, Yigang},
  booktitle={Proceedings of the 33rd ACM International Conference on Multimedia},
  pages={4914--4922},
  year={2025}
}

@article{wah2011caltech,
  title={The caltech-ucsd birds-200-2011 dataset},
  author={Wah, Catherine and Branson, Steve and Welinder, Peter and Perona, Pietro and Belongie, Serge},
  year={2011}
}

@article{wang2026data,
  title={Data-Free Class-Incremental Gesture Recognition With Prototype-Guided Pseudo-Feature Replay},
  author={Wang, Hongsong and Sun, Ao and Gui, Jie and Wang, Liang},
  journal={IEEE Transactions on Image Processing},
  volume={35},
  pages={3623--3632},
  year={2026},
  publisher={IEEE}
}
\end{document}

% ===============================================================
% 标题与盲审设置区
% ===============================================================
\title{Virtual Category-Guided Continual Generalized Category Discovery\\-\textit{Supplementary Material}-}
\titlerunning{Supplementary Material}

% 盲审隐藏作者

\author{Jiahui Xiong\inst{1} \and
Qiuxia Lai\inst{2}\textsuperscript{*} \and
Hongsong Wang\inst{1,3}\textsuperscript{*}}
\authorrunning{J.~Xiong et al.}

\institute{
$^1$School of Computer Science and Engineering, Southeast University, Nanjing 210096, China \\
$^2$State Key Laboratory of Media Convergence and Communication, Communication University of China, Beijing 100024, China \\
$^3$Key Laboratory of New Generation Artificial Intelligence Technology and Its Interdisciplinary Applications (Southeast University), Ministry of Education, China \\
\email{\{hongsongwang, jiahuixiong\}@seu.edu.cn, qxlai@cuc.edu.cn}
}

\renewcommand{\thefootnote}{\ensuremath{*}}
\footnotetext[1]{Corresponding authors.}

\maketitle

% ===============================================================
% 附录编号重置区 (关键！)
% 将编号改成 A, B, Figure A1, Table A1
% ===============================================================
\appendix
\renewcommand{\thesection}{\Alph{section}}
\renewcommand{\thefigure}{\Alph{section}\arabic{figure}}
\renewcommand{\thetable}{\Alph{section}\arabic{table}}
\renewcommand{\theequation}{\Alph{section}\arabic{equation}}
\setcounter{section}{0}
\setcounter{figure}{0}
\setcounter{table}{0}
\setcounter{equation}{0}

% ===============================================================
% 附录正文区
% ===============================================================
We provide this supplementary material to complement and further support the main paper. This document presents additional details regarding our proposed method and experimental settings. Specifically, we provide additional experimental details in Sec. \ref{sec:details}, including the dataset split protocol (Sec.\ref{sec:dataset}) and the evaluation metrics used in continual generalized category discovery (Sec.\ref{sec:metrics}). Then, we first present the pseudo-code of our proposed framework in Sec.\ref{sec:pseudo-code} to clarify the training procedure during the online continual learning stage. In Sec.\ref{sec:tsne}, we present additional t-SNE visualizations to illustrate the representation quality learned by our method across continual sessions. Section.\ref{sec:more_results} collects additional experimental results including broader comparisons, evaluations on more datasets, sensitivity studies, generality verification, standard deviation statistics, and component-wise ablation analysis. Finally, Sec.\ref{sec:impacts} discusses the potential impacts of this work.

\section{More Experiment Details}
\label{sec:details}

\subsection{Dataset Details}
\label{sec:dataset}
Following standard C-GCD protocols, experiments are conducted on CIFAR-100~\cite{krizhevsky2009learning}, ImageNet-100~\cite{deng2009imagenet}, and Tiny-ImageNet~\cite{li2023imbagcd}. Each dataset is divided into an offline initialization stage and multiple online continual sessions, as we show in \ref{tab: split}. In the offline stage, the model is trained with labeled data from 50\% of the categories, where each selected class contributes 80\% of its training samples to form the initial known set $C_{\text{init}}$. In the online stage, the remaining 50\% of categories are introduced sequentially as unlabeled data streams. Each session contains a mixture of previously seen classes and newly emerging categories. Evaluation is conducted after each session on test sets containing both old and newly discovered classes.

\begin{table}[htbp]
  \label{tab: split}
  \centering
  \caption{Dataset splits of C-GCD setting. The data configuration for the scenario with 50\% categories labeled and 5 continual sessions.}
  \resizebox{\linewidth}{!}{
    \begin{tabular}{c|c|c|c|c|c}
    \toprule
    \multirow{2}[4]{*}{Datasets} & \multicolumn{2}{c|}{Initial Session} & \multicolumn{3}{c}{Continual Sessions} \\
\cmidrule{2-6}          & categories & samples per category & novel categories & samples per novel & samples per old \\
    \midrule
    CIFAR-100 & 50    & 400   & 10    & 400   & 25 \\
    TinyImageNet & 100   & 400   & 20    & 400   & 25 \\
    ImageNet-100 & 50    & 1000  & 10    & 1000  & 60 \\
    \bottomrule
    \end{tabular}%
  }
  \label{tab:addlabel}%
\end{table}

\subsection{Metrics Details}
\label{sec:metrics}
Following conventional protocol, we evaluate continual behavior using the forgetting metric $M_f$ and the discovery metric $M_d$. 

The forgetting metric is defined as
\begin{equation}
    {M_f} = ACC_\text{known}^0 - \mathop {\min }\limits_{1 \le t \le T} \{ ACC_\text{known}^t\},
\end{equation}
which quantifies the maximum degradation of previously learned categories during continual learning. In our work, $M_f$ is reported using explicit session-wise accuracies to provide a clearer and more transparent comparison across methods.

The discovery metric is defined as
\begin{equation}
    {M_d} = \frac{1}{T}\sum\limits_{1 \le t \le T} {ACC_\text{noval}^t},
\end{equation}
which measures the average recognition accuracy over all encountered categories across sessions and reflects the model’s sustained capability to discover and integrate novel classes. This definition is consistent with previous studies to ensure a fair comparison of continual discovery performance.

\section{Pseudo-code of our Algorithm}
\label{sec:pseudo-code}
In this section, we present a detailed pseudo code of our work in Algorithm \ref{alg:vcgcd}.

\vspace{-0.5em}
\begin{algorithm}[H]
\caption{Pseudo-code of Virtual Category-Guided C-GCD in a PyTorch-like style.}
\label{alg:vcgcd}

\begin{lstlisting}[
    language=Python,
    basicstyle=\ttfamily\small,
    commentstyle=\color{green!50!black}\ttfamily,
    keywordstyle=\color{blue}\bfseries,
    stringstyle=\color{purple},
    breaklines=true,
    columns=flexible,
    frame=none,
    xleftmargin=0pt,
    aboveskip=0pt,
    belowskip=0pt,
    upquote=true,        % 修复引号编码问题
    mathescape=false     % 防止$被解析为数学模式
]
# Input:  backbone encoder f_theta, classifier g_phi, unlabeled data, 
#    feature bank F, hyperparameters (lambda_1, lambda_2, num_neighbors)
# Output: trained student model (composition of encoder and classifier)

# ==================== Model Initialization ====================
student = nn.Sequential(encoder, classifier)
teacher = copy.deepcopy(student)              # EMA teacher initialization
feature_bank = FeatureBank()                  # Initialize memory buffer
optimizer = torch.optim.Adam(student.parameters(), lr=1e-3)

# ==================== Continual Learning Loop ====================
for session in sessions:
    dataloader = DataLoader(session, batch_size=B, shuffle=True)    
    for epoch in range(E):
        for x in dataloader:                  # x: unlabeled batch (B, C, H, W)
            # Dual-view augmentation
            x_w = weak_augment(x)             # Weak aug for teacher
            x_s = strong_augment(x)           # Strong aug for student   
            # Teacher inference (no gradient)
            with torch.no_grad():
                f_t = teacher[0](x_w)         # Teacher features
                logits_t = teacher[1](f_t)    # Teacher predictions
                pseudo_labels = logits_t.argmax(dim=-1)    
            # Student feature extraction
            f_s = encoder(x_s)
            PC = top_k_candidates(logits_t, k=top_k)  # Potential categories            
            # Virtual Category Learning
            logits_s = classifier(f_s)
            l_v = compute_virtual_logit(f_s, PC)      # Virtual logit
            logits_ext = torch.cat([logits_s, l_v.unsqueeze(-1)], dim=-1)
            loss_vc = vc_loss(logits_ext, PC)            
            # Expanded Neighborhood Contrastive Learning
            N = knn(f_s, feature_bank, k=K)           # K-nearest neighbors
            EM = union_neighbors(N)                   # Expanded neighborhood set
            loss_encl = contrastive_loss(f_s, N, EM)
            
            # Total objective and optimization
            loss_base = baseline_loss(logits_s, pseudo_labels)
            loss = loss_vc + lambda_1 * loss_encl + lambda_2 * loss_base            
            optimizer.zero_grad()
            loss.backward()
            optimizer.step()           
            # Teacher EMA update and feature bank update
            teacher = ema_update(student, teacher, momentum=alpha)
            feature_bank.update(f_s.detach())

return trained student model
\end{lstlisting}

\end{algorithm}

\section{Visualization Representation}
\label{sec:tsne}

\begin{figure*}[t]
\centering
\includegraphics[width=\textwidth]{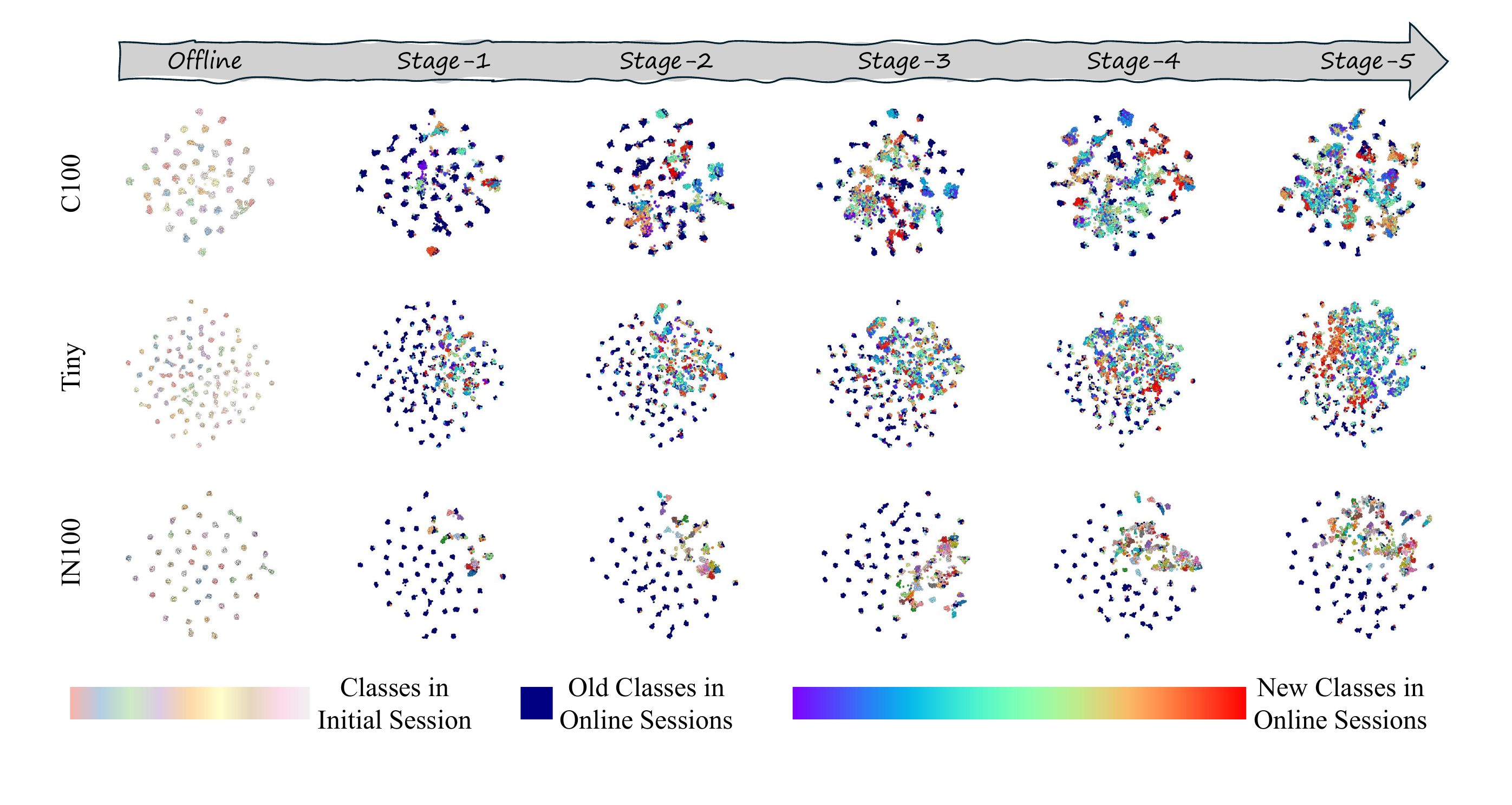}
\caption{TSNE visualization of CIFAR-100, Tiny-ImageNet, and ImageNet-100 datasets with features from our work at each stage. See Sec.\ref{sec:tsne} for details.}
\label{tsne}
\end{figure*}

We provide additional qualitative visualizations to illustrate the effectiveness of our proposed method in learning discriminative representations for both known and novel categories across continual sessions.
We employ t-SNE to project the high-dimensional features extracted from the penultimate layer of our model into a 2D space. We visualize the feature distributions on CIFAR100, ImageNet-100 and Tiny-ImageNet datasets across all continual stages.

As shown in Fig.\ref{tsne}, we adopt the following color coding scheme: (1) for the Offline stage, where only known categories exist, we use the Pastel1 qualitative colormap to distinguish different classes; (2) for Online stages, known categories (including previously discovered classes) are uniformly rendered in navy blue to emphasize the model's retention capability, while novel categories discovered at each session are visualized using the tab20 qualitative colormap with vibrant colors to highlight their discriminability.

It is observed that our method maintains well-separated clusters for known categories (navy dots) while effectively grouping novel categories into distinct, color-separated clusters across sessions. Notably, the visualization demonstrates that our Virtual Category Learning mechanism successfully prevents representation collapse, as novel classes remain visually separable from known classes despite the absence of labels during continual learning. Furthermore, the compactness of clusters indicates that our Expanded Neighborhood Contrastive Learning enhances feature discriminability, facilitating reliable pseudo-labeling in subsequent stages.

\section{More Results}
\label{sec:more_results}
\subsection{Broader Comparisons} We have added comparisons with two recent 2025 baselines ~\cite{zhao2025multi,hao2025tree}. For fair comparison, we follow their described settings and report the average over all sessions. Table \ref{tab:broader_cmp} shows our superior performance. 

\begin{table}[h]    
  \centering
  \caption{Comparison with recent baselines on C100, Tiny and IN100.}
  \label{tab:broader_cmp}
  \vspace{-10pt}
    \setlength\tabcolsep{3pt}
    \renewcommand\arraystretch{1.0}
    \resizebox{0.7\linewidth}{!}{
      \begin{tabular}{c|ccc|ccc|ccc}
        \hline\noalign{\hrule height 0.1pt}
        \multirow{2}{*}{Methods} & \multicolumn{3}{c|}{C100} & \multicolumn{3}{c|}{Tiny} & \multicolumn{3}{c}{IN100} \\
        \cline{2-10}
              & All   & Old   & New   & All   & Old   & New   & All   & Old   & New \\
        \hline
        MCMI [1]  & 61.30  & 67.00  & 59.00  & 59.50  & 59.70  & 58.90  &-       &-       &-  \\
        ToP [2]   & 66.90  & 74.47  & 68.50  &-       &-       &-       & 78.33  & 82.57  & 78.90 \\
        % DMCL\cite{jin2026dual}  & 76.36  & 77.89  & \textbf{70.01} & \textbf{72.82} & 75.52  & 64.70  & 81.83  & 83.17  & 78.41  \\
        Ours  & \textbf{76.94} & \textbf{78.10} & 68.20  & \textbf{68.40}  & \textbf{77.38} & 57.63 & \textbf{84.43} & \textbf{87.40} & 66.13 \\
        \hline
      \end{tabular}
    }
    \vspace{-10pt}
\end{table}

\subsection{More Datasets}
We further evaluate on a fine-grained dataset CUB~\cite{wah2011caltech}. Table \ref{tab:cub} shows that our method outperforms the other two strong baselines ~\cite{wu2023metagcd,ma2024happy}. 
\begin{table}[h]
    \vspace{-18pt}
  \centering
  \caption{Session-wise accuracy on the fine-grained CUB dataset.}
  \label{tab:cub}
  \setlength\tabcolsep{1.2pt}
  \renewcommand\arraystretch{1.0}
  \resizebox{\linewidth}{!}{%
  \begin{tabular}{l|c|ccc|ccc|ccc|ccc|ccc}
    \hline\noalign{\hrule height 0.1pt}
    Methods & Offline & \multicolumn{3}{c|}{Session-1} & \multicolumn{3}{c|}{Session-2} & \multicolumn{3}{c|}{Session-3} & \multicolumn{3}{c|}{Session-4} & \multicolumn{3}{c}{Session-5} \\
    \cline{2-17}
          & All   & All   & Old   & New   & All   & Old   & New   & All   & Old   & New   & All   & Old   & New   & All   & Old   & New \\
    \hline
    MetaGCD & 89.20  & 67.08  & 70.21  & 51.92  & 60.77  & 62.39  & 50.86  & 57.53  & 59.33  & 37.78  & 51.90  & 52.22  & 49.40  & 49.60  & 49.96  & 46.38  \\
    % \hline
    Happy & 90.26  & 79.51  & 86.96  & 43.36  & 71.23  & 84.78  & 37.74  & 65.28  & 80.51  & 40.03  & 59.33  & 75.42  & 39.38  & 58.53  & 77.05  & 40.17 \\
    % \hline
    Ours  & 90.11  & \textbf{79.99}  & \textbf{87.34}  & \textbf{44.37}  & \textbf{72.15}  & 83.91  & \textbf{43.05}  & \textbf{67.31}  & \textbf{82.77}  & \textbf{41.70}  & \textbf{62.94}  & \textbf{79.99}  & \textbf{41.79}  & \textbf{58.75}  & \textbf{78.29}  & 39.38  \\
    \hline
    \end{tabular}%
  }
  \vspace{-16pt}
\end{table}

\subsection{Generality Beyond a Single Baseline} We further integrated VCL and ENCL into another baseline, MetaGCD. As shown in Table~\ref{tab:metagcd_as_baseline}, ``+VCL+ENCL'' consistently improves over MetaGCD on C100/Tiny, supporting the generality. 
\begin{table}[h]
  \centering
  \caption{Generality verification by integrating VCL and ENCL into MetaGCD on C100 and Tiny-ImageNet.}
  \label{tab:metagcd_as_baseline}
  \vspace{-10pt}
    \setlength\tabcolsep{1.2pt}
    \renewcommand\arraystretch{1.0}
  \resizebox{\linewidth}{!}{%
  \begin{tabular}{c|l|c|ccc|ccc|ccc|ccc|ccc}
    \hline\noalign{\hrule height 0.1pt}
    \multirow{2}{*}{Datasets} 
    & \multirow{2}{*}{Methods} 
    & Offline 
    & \multicolumn{3}{c|}{Session-1} 
    & \multicolumn{3}{c|}{Session-2} 
    & \multicolumn{3}{c|}{Session-3} 
    & \multicolumn{3}{c|}{Session-4} 
    & \multicolumn{3}{c}{Session-5} \\
    \cline{3-18}
    & 
    & All & All & Old & New 
    & All & Old & New 
    & All & Old & New 
    & All & Old & New 
    & All & Old & New \\
    \hline
    \multirow{2}{*}{C100} & MetaGCD & 90.82 & 76.12 & 83.60 & 38.70 & 69.40 & 72.82 & 48.90 & 61.95 & 65.76 & 35.30 & 58.22 & 61.21 & 34.30 & 55.78 & 58.47 & 31.60 \\
    % \cline{2-2}
    & +VCL+ENCL & 90.79 & \textbf{78.39} & 82.31 & \textbf{58.74} & \textbf{71.33} & \textbf{78.64} & \textbf{52.68} & \textbf{63.86} & \textbf{73.07} & \textbf{39.43} & \textbf{60.19} & \textbf{65.97} & \textbf{35.82} & \textbf{57.39} & \textbf{58.63} & \textbf{33.55} \\
    \hline
    \multirow{2}{*}{Tiny} & MetaGCD & 84.20 & 60.88 & 64.90 & 40.80 & 57.20 & 61.03 & 34.20 & 54.36 & 57.19 & 34.60 & 50.83 & 53.59 & 28.80 & 48.14 & 50.16 & 30.00 \\
    % \cline{2-2}
    & +VCL+ENCL & 84.37 & \textbf{62.64} & 64.23 & \textbf{53.11} & \textbf{59.04} & \textbf{65.42} & \textbf{36.91} & \textbf{55.96} & \textbf{62.46} & \textbf{37.09} & \textbf{53.33} & \textbf{62.50} & 28.34 & \textbf{49.97} & \textbf{51.08} & \textbf{30.91} \\
    \hline
  \end{tabular}%
  }
  \vspace{-12pt}
\end{table}

\subsection{Standard Deviations} Performance comparisons with \texttt{std}s on C100 are summarized in Table~\ref{tab:std}, which show consistent gains over Happy ~\cite{ma2024happy}. All statistics are collected from multiple independent runs to eliminate random training fluctuations. The small deviation ranges further demonstrate that our method maintains stable optimization dynamics across continual sessions.

\begin{table}[h]
  \centering
  \caption{Performance comparison with standard deviations on C100.}
  \label{tab:std}
  \vspace{-10pt}
    \setlength\tabcolsep{0.8pt}
    \renewcommand\arraystretch{1.2}
  \resizebox{\linewidth}{!}{%
  \begin{tabular}{c|ccc|ccc|ccc|ccc|ccc}
    \hline\noalign{\hrule height 0.1pt}
    \multirow{2}{*}{} 
    & \multicolumn{3}{c|}{Session-1} 
    & \multicolumn{3}{c|}{Session-2} 
    & \multicolumn{3}{c|}{Session-3} 
    & \multicolumn{3}{c|}{Session-4} 
    & \multicolumn{3}{c}{Session-5} \\
    \cline{2-16}
    & All & Old & New 
    & All & Old & New 
    & All & Old & New 
    & All & Old & New 
    & All & Old & New \\
    \hline
    Happy & $80.04$ & $85.26$ & $56.10$ 
          & $74.13$ & $78.27$ & $49.30$ 
          & $68.23$ & $70.86$ & $49.80$ 
          & $62.26$ & $63.75$ & $50.30$ 
          & $59.99$ & $60.96$ & $51.30$ \\
    \hline
    Ours & $\textbf{83.14}_{\pm0.06}$ & $\textbf{83.84}_{\pm0.05}$ & $\textbf{79.63}_{\pm0.54}$ 
          & $\textbf{76.30}_{\pm0.07}$ & $\textbf{83.62}_{\pm0.82}$ & $\textbf{57.98}_{\pm2.02}$ 
          & $\textbf{70.16}_{\pm0.15}$ & $\textbf{79.46}_{\pm0.78}$ & $\textbf{54.54}_{\pm1.61}$ 
          & $\textbf{63.56}_{\pm0.50}$ & $\textbf{80.49}_{\pm1.29}$ & $\textbf{}42.39_{\pm2.54}$ 
          & $\textbf{61.25}_{\pm0.61}$ & $\textbf{62.41}_{\pm0.70}$ & $\textbf{}48.09_{\pm0.73}$ \\
    \hline
  \end{tabular}%
  }
  \vspace{-10pt}
\end{table}

\subsection{Component-Wise Ablation}  We have conducted additional ablation experiments to analyze VCL/ENCL effects on old-class retention ($M_f$) and novel-class discovery ($M_d$), as reported in Table~\ref{tab:component_ab_study}. 
\begin{table}[h]
  \centering
  \caption{Ablation on forgetting metric \(M_f\) and discovery metric \(M_d\).}
  \label{tab:component_ab_study}
  \vspace{-10pt}
    \setlength\tabcolsep{7.0pt}
    \renewcommand\arraystretch{0.9}
  \resizebox{0.5\linewidth}{!}{%
  \begin{tabular}{cc|cc|cc}
    \hline\noalign{\hrule height 0.1pt}
          &       & \multicolumn{2}{c|}{C100} & \multicolumn{2}{c}{Tiny} \\
    \hline
    VC    & EN & $M_f~(\downarrow)$    & $M_d~(\uparrow)$    & $M_f~(\downarrow)$    & $M_d~(\uparrow)$ \\
    \hline
    ×     & ×     & 29.40  & 51.36  & 8.66  & 35.23  \\
    \checkmark      & ×     & 29.94  & 52.43  & 8.55  & 35.67  \\
    ×     & \checkmark      & 28.53  & 53.10  & 9.31  & 39.22  \\
    \checkmark      & \checkmark      & \textbf{27.94}  & \textbf{56.61}  & 9.17  & \textbf{40.56}  \\
    \hline
  \end{tabular}%
  }
\end{table}

\section{Potential Impacts}
\label{sec:impacts}
\subsection{Limitations}
While our method demonstrates robust discovery performance under standard continual learning protocols, several challenges remain to be fully addressed. First, accurately estimating the number of emerging categories in highly dynamic data streams remains difficult, particularly when novel classes appear gradually or with severe imbalance, as the quality of initial clustering can significantly impact subsequent learning. Second, although Virtual Category Learning mitigates confirmation bias by handling ambiguous samples, pseudo-labeling strategies remain sensitive to noisy predictions in early sessions, leading to potential error accumulation over extended learning horizons. Finally, while our framework maintains representation stability under moderate distribution changes, modeling fine-grained categories with limited samples and handling strong distribution shifts over long-term continual streams require further investigation. These challenges represent open problems that the broader C-GCD community continues to address.

\subsection{Future Work}
VCL and ENCL are objective-level components potentially compatible with replay strategies and other continual learning regularizers. By adapting the PC set construction and neighborhood sampling rules, our framework can be integrated with such methods to further balance stability and plasticity. We leave systematic investigation of these combinations to future work.

\subsection{Social Impacts}
This work studies C-GCD, which aims to enable visual recognition systems to continuously discover new categories from unlabeled data streams while maintaining knowledge of previously learned classes. Such capability can benefit a wide range of real-world applications where the visual environment evolves over time. For instance, autonomous systems, robotics, and large-scale monitoring platforms often encounter previously unseen objects or concepts after deployment. By allowing models to incrementally recognize and organize emerging categories without retraining from scratch, our approach can improve the adaptability and scalability of long-term learning systems.

\bibliographystyle{splncs04}
\bibliography{main}